\documentclass[12pt]{article}

\long\def\nop#1{}
\def\proof{\noindent {\sl Proof.\ \ }}
\def\qed{\hfill{\boxit{}}}
\long\def\boxit#1{\vbox{\hrule\hbox{\vrule\kern3pt
                  \vbox{\kern3pt#1\kern3pt}\kern3pt\vrule}\hrule}}

\def\Z{{\cal Z}}
\let\plural=\relax
\def\color[#1]#2{}
\def\mod{M\!od}
\def\true{{\sf true}}
\def\false{{\sf false}}
\def\l{\langle}
\def\r{\rangle}

\def\modifymargins#1#2{
\newdimen\addtoh
\newdimen\addtow
\addtoh=#1
\addtow=#2

\advance\topmargin by -\addtoh
\multiply\addtoh by 2
\advance\textheight by \addtoh

\advance\oddsidemargin by -\addtow
\advance\evensidemargin by -\addtow
\multiply\addtow by 2
\advance\textwidth by \addtow
}
\expandafter\ifx\csname ttytty\endcsname\relax\modifymargins{40pt}{30pt}\fi

\def\con{\mathrm con}
\def\maxcon{\mathrm maxcon}
\def\subsat{\mathrm subsat}

\def\possnewtheorem#1#2{
\expandafter\ifx\csname #1\endcsname\relax
\newtheorem{#1}{#2}
\fi
}

\possnewtheorem{theorem}{Theorem}
\possnewtheorem{corollary}{Corollary}
\possnewtheorem{lemma}{Lemma}
\possnewtheorem{definition}{Definition}
\possnewtheorem{question}{Question}
\possnewtheorem{example}{Example}
\possnewtheorem{counterexample}{Counterexample}

\def\ttytex#1{#1\nop}			

\title{Merging with unknown reliability}
\author{Paolo Liberatore}

\begin{document}

\maketitle

\begin{abstract}

Merging beliefs depends on the relative reliability of their sources. When
unknown, assuming equal reliability is unwarranted. The solution proposed in
this article is that every reliability profile is possible, and only what holds
according to all is accepted. Alternatively, one source is completely reliable,
but which one is unknown. These two cases motivate two existing forms of
merging: maxcons-based merging and arbitration.

\end{abstract}

\section{Introduction}

Most of the literature on belief merging concerns sources of the information of
equal reliability~%
\cite{libe-scha-98-b,lin-mend-99,chop-etal-06,ever-etal-10,koni-pere-11}.
Such a scenario occurs, but not especially often. Two identical temperature
sensors produce readings that are equally likely to be close to the actual
value, but a difference in made, age, or position changes their reliability.
Two experts hardly have the very same knowledge, experience and ability. The
reliability of two databases on a certain area may depend on factors that are
unknown when merging them.

Merging under equal and unequal reliability are two scenarios, but a third
exists: unknown reliability. Most previous work in belief merging is about the
first~\cite{libe-scha-98-b,lin-mend-99,chop-etal-06,ever-etal-10,koni-pere-11,%
hare-etal-20,ever-etal-20}; some is about the second~%
\cite{reve-97,lin-96,chol-98,koni-lang-marq-04}; this one is about the third.

The difference between equal and unknown reliability is clear when its
implications on some examples are shown.

\

\begin{tabular}{lll}
example	&
equal reliability &
unknown reliability
\\
\hline
two experts &
\vbox{\hsize 5cm\noindent
they have the very same knowledge, experience and ability} &
\vbox{\hsize 5cm\noindent
they differ in knowledge, experience and ability,
but how much of these they posses is unknown}
\\
\hline
two sensors &
\vbox{\hsize 5cm\noindent
they are of the same kind and are in the same condition
(example, temperature sensors located next to each other,
distance sensors with the same orientation)} &
\vbox{\hsize 5cm\noindent
they are of different kind, or are in different conditions,
and which one is more reliable in the current situation is not known}
\\
\hline
two databases &
\vbox{\hsize 5cm\noindent
they cover the very same domain, and are equally likely to be correct} &
\vbox{\hsize 5cm\noindent
they cover different domains, so that a certain piece of information may have
been crucial to one but a detail in the second}
\\
\hline
\end{tabular}

\

The assumption of equal reliability is quite strong in the example of the two
experts; rather, there may be some reason to believe one more than the other;
not knowing who, this scenario falls in the case of unknown reliability. For
the two sensors and the two databases equal reliability is not unlikely, but so
is the case of unknown reliability.

If reliability is unknown, can it be assumed equal?

When merging preferences, yes. When merging beliefs, no.

Merging preferences~\cite{list-13,lang-04,mata-pere-17} aims at obtaining a
result that best reflects the collective opinion of a group. A common premise
is that all members of the group should have the same weight on the final
decision, as formalized by the condition of anonymity. In lack of information
telling otherwise, equal weights are a valid assumption.



A technical example shows why not when merging beliefs instead. Three scenarios
are possible: $A$, $B$ and $C$; the two sources of information rank they
unlikeliness on a scale from $0$ to $3$, with $0$ being the most likely and $3$
the least (unlikeliness scales are common in belief
revision~\cite{kats-mend-91-b,darw-pear-97,rott-06a}, in spite of likeliness
being more intuitive). The first source grades $A$ as a most unlikely scenario,
the second as a most likely; numerically, these are unlikeliness $3$ and $0$.
Both sources grade $B$ as kind of likely ($1$), and $C$ in the opposite way of
$A$ ($0$ and $3$).

\

\ttytex{
\begin{center}
\begin{tabular}{lll}
{scenario} &
{unlikeliness according} &
{unlikeliness according} \\
&
{to the first source} &
{to the second source} \\
\hline
$A$ & 3 & 0 \\
$B$ & 1 & 1 \\
$C$ & 0 & 3
\end{tabular}
\end{center}
}{
scenario    unlikeliness according    unlikeliness according
            to the first source       to the second source
A           3                         0
B           1                         1
C           0                         3
}

\

The relative reliability of the sources is unknown. Assuming them equal, the
result is that $B$ is the most likely scenario, since its overall unlikeliness
is minimal: $1+1$ is less than $3+0$ and $0+3$.

A further piece of information then arrives: the first source is twice as
reliable as the second. Doubling the numbers coming from the first source
changes the plausibility of the scenarios:

\

\ttytex{
\begin{center}
\begin{tabular}{lll}
{scenario} &
{unlikeliness according} &
{unlikeliness according} \\
&
{to the first source, weighted} &
{to the second source, weighed} \\
\hline
$A$ & $2 \times 3$ & $0$ \\
$B$ & $2 \times 1$ & $1$ \\
$C$ & $2 \times 0$ & $3$
\end{tabular}
\end{center}
}{
scenario    unlikeliness according              unlikeliness according
            to the first source, weighted       to the second source, weighted
A           2 * 3                               0
B           2 * 1                               1
C           2 * 0                               3
}

\

Scenario $C$ now has the same unlikeliness of $B$ ($2 \times 0 + 3 = 2 \times 1
+ 1 = 3$) and both have less than $A$ ($2 \times 3 + 0 = 6$).

Beforehand, merging selected scenario $B$ only. After, it was either $B$ or
$C$. Adding information makes the result less precise: before, merging uniquely
identified a single scenario; after, it is undecided between two.

This happened because of the addition of some information: the first source is
more reliable than the second. This new information is totally consistent with
what known before, since what was known before was that the relative
reliability of the sources is unknown, meaning that the first could be less,
equally or more reliable than the second. This is not a shift of information,
only an addition. Yet, it leads to a loss of information.



This example is inspired by the ``penny $z$'' of
Popper~\cite[pp.~425--426]{popp-59}: a coin is initially assumed fair in lack
of information indicating otherwise; adding the confirmation of fairness does
not change its probability of falling heads or tails. In the interpretation of
probability as degree of belief~\cite{haje-12}, the probabilities are the
epistemic state. Adding information should alter the epistemic state, but the
addition of fairness (which is new information) changes nothing.

This example was used by Popper against the subjective interpretation of
probabilities, but relies on the principle of indifference: events of unknown
probability are assumed equally probable~\cite{keyn-21,shac-07}. The Bertrand
paradox~\cite{bert-89,keyn-21,shac-07} shows it is problematic; the coin
example by Popper~\cite{popp-59} shows another contradictory aspect of
it~\cite{gard-sahl-82}.


The belief merging version of the principle of indifference is the assumption
of equal reliability in lack of information about the relative reliability of
the sources. In the subjective interpretation of probability, the probability
of an event is the degree of belief in that event happening~\cite{haje-12}; in
belief merging, the weight of a source is the likeliness of the formulae it
provides being true, or at least close to
truth~\cite{reve-97,lin-96,chol-98,koni-lang-marq-04,darw-marq-04}. The event
``formula $F$ is true in the real world'' provides a qualitative connection of
probability with merging. The principle of indifference translates into the
assumption of equal reliability.


The probability version of sources of unknown reliability is the lack of
knowledge of the probability of events. Economists distinguish between risk
(known probability) and Knightian uncertainty (unknown
probability)~\cite{nish-ozak-07}. An often-used example is the urn containing
twenty yellow balls and forty balls of another color, which may be either blue
or green; these forty are either all blue or all green, but which of the two is
not known. This scenario involves both risk (the probability of yellow or not
yellow is known) and Knightian uncertainty (the presence of blue balls is
unknown).

This urn suggests a way to deal with the problem in belief merging. The
probability of drawing a yellow ball is always one third, but assuming the same
for blue and green is as if the urn contained twenty balls for each color. This
may be acceptable for a single drawn, but an example shows it is not in
general. The probability of drawing two balls of the same color (putting the
first ball back in the urn) under the assumption of equal probability is
{} $\frac{1}{3}$
instead of
{} $\frac{1}{3} \times \frac{1}{3} + \frac{2}{3} \times \frac{2}{3}$.
The first value is obtained by selecting from the nine possible outcomes of
probability $\frac19$ each (random first ball and random second ball) only the
three where the balls are the same color: $3 \times \frac19 = \frac13$. The
second value can be obtained by considering the second drawn not independent of
the first, but also by calculating the probability under the assumption of
forty blue balls: the probability of the two balls being both yellow is
$\frac13 \times \frac13$, that of being both blue is $\frac23 \times \frac23$.
Importantly, the very same value is obtained for forty green balls instead. Not
only this probability holds in both cases, it resists the addition of
information. It holds even if it is later discovered that the urn is made in a
factory that normally uses forty green balls, and the blue ball version is a
rare collector's edition.


In terms of belief merging, two sources of unknown reliability may have the
same reliability, or one may be more reliable than the other. All cases are
considered, and only what holds in all of them is taken. This is analogous to
reasoning from multiple probability distributions~\cite{halp-tutt-93}.

%
%
%

How does this solution work in the example of the three scenarios $A$, $B$ and
$C$ respectively ranked $[3,0]$, $[1,1]$ and $[0,3]$? The first is preferred
when the second source is much more reliable than the first, the second when
they are equally reliable and the last when the first is much more reliable
than the first. When reliability was unknown, none of the three scenarios was
considered more likely than the others. When the first source is discovered to
be much more reliable than the second, likeliness of scenarios $A$ decreases.
An addition of information (the relative reliability of the sources) leads to
an increase of information (from all three scenarios to only $B$ and $C$).

Such a likeliness evaluation is not that easy when reliability is encoded
numerically rather than qualitatively. The first source may be $10\%$ more
likely than the second, still making $B$ the only most likely scenario. A
potentially infinite number of reliability indicators are involved. However, it
will be proved that in all cases only a finite number of them need to be
considered.

Another result in this article is that the disjunction of all maxcons~%
\cite{brew-89,bara-etal-92,benf-etal-97,koni-pere-11,ammo-etal-15,%
gran-hunt-11,dubo-etal-16} is the result of merging formulae of unknown
reliability using the drastic distance. This result invalidates the view that
maxcons are unsuitable for merging since they do not take into account the
distribution of information~\cite{koni-00,koni-pere-11}. Rather, they do
exactly what they should; the problem was with the assumption of equal
reliability.

Technically, merging is defined by selecting the models at a minimal weighted
distance from the formulae provided by the sources. The drastic and the Hamming
distances are considered as two relevant examples. This is detailed in the next
section. The following proceed by formalizing the idea of this article: weights
represent the unknown reliability, so they may range arbitrarily; all models
obtained from some possible weights are selected. The results obtained in the
following sections are:

\begin{itemize}

\item merging with arbitrary weights is related to a direct comparison of
models; the result is the same when the distance function has binary codomain,
but not in the general case;

\item the drastic distance has binary codomain; therefore, merging can be done
by a direct comparison of models; a consequence is that it is the same as
disjoining the maxcons;

\item the Hamming distance allows for a general existence result: all distances
can be obtained from suitable formulae; in particular, some formulae require an
exponential number of weights;

\item while the definition of merging with unknown reliability involves a
universal quantification over weights, only an exponential number of these are
really required; also, selection of a model can be checked by comparing it only
with $m$ other models at time, where $m$ is the number of models to be checked;
when merging two formulae, this result allows for a graphical representation of
merging as the part of the convex hull of a set of points that is visible from
the origin;

\item merging with unknown reliability satisfy most of the postulate by
Konieczny and Perez~\cite{koni-pere-11}, including the original version of the
arbitration postulate~\cite{koni-pere-98,meye-01};

\item with a suitable restriction of the possible weights arbitration by
closest pairs of models~\cite{libe-scha-98-b} is recovered.

\end{itemize}

The second-last section of the article briefly considers the case of sources
providing more than one formula. The last discusses the results obtained in
this article.

\section{Merge by weights}

In this article, merging is done by minimizing the weighting distance of models
obeying integrity constraints from the formulae to be merged. The integrity
constraints are denoted $\mu$, the formulae to be merged $F_1,\ldots,F_m$. This
is the basic settings for belief merge, where each source provides exactly one
formula $F_i$; the case of multiple formulae is considered in a following
section.

Models are represented by the set of literals they satisfy; for example
$I=\{a,\neg b,c\}$ is the model assigning false to $b$ and true to $a$ and $c$.
The distance between two models $I$ and $J$ is an integer denoted by $d(I,J)$.
Two intuitive and commonly used distances are the drastic and the Hamming
distance. The drastic distance is defined by $dd(I,J) = 0$ if $I=J$ and
$dd(I,J)=1$ otherwise. The Hamming distance $dh(I,J)$ is the number of literals
assigned different truth values by $I$ and $J$; for example,
{} $dh(\{a,\neg b,c\},\{\neg a,\neg b,\neg c\})=2$,
since the two models differ on $a$ and $c$. Other distances can be defined;
they are assumed to satisfy $d(I,I)=0$ and $d(I,J)>0$ if $I \not= J$.

Distance extends from models to formulae: regardless of which distance is used,
$d(I,F)$ is the minimal value of $d(I,J)$ for every $J \models F$. It further
extends from a formula to a list of them: the distance between a model and a
list of formulae is the array of integers
{} $d(I,F_1,\ldots,F_m) = [d(I,F_1),\ldots,d(I,F_m)]$.

Merging by weighed distance has been the historically first way of integrating
formulae coming from sources of different reliability. Given a vector of
positive integers $W = [w_1,\ldots,w_m]$, the weighted distance of a model $I$
from the formulae is $W \cdot d(I,F_1,\ldots,F_m)$, where the dot stands as
usual for the scalar product:

\[
[w_1,\ldots,w_m] \cdot d(I,F_1,\ldots,F_m) =
\sum_{1 \leq i \leq m} w_i \times d(I,F_i)
\]

This product defines a single integer telling the aggregated distance from $I$
to the formulae $F_i$, weighted by the relative reliability of each as
represented by the integer $w_i$. Merging selects the models satisfying the
integrity constraints $\mu$ that have minimal weighted distance from the
formulae.

$$
\Delta^{d,W,\cdot}_\mu(F_1,\ldots,F_m) =
\{I \models \mu \mid W \cdot d(F_1,\ldots,F_m) \mbox{ is minimal } \}
$$

This function depends on three parameters: a model-to-model distance $d$, a
vector of weights $W=[w_1,\ldots,w_m]$ and a mechanism for distilling a single
amount out of these and $d(I,F_1,\ldots,F_m)$, in this case the scalar product.

\

Fixed weights are used when the relative reliability of the sources is known.
Weights $W=[1,\ldots,1]$ makes the scalar product the same as a sum, and
weighted merge the same as the usual operators based on the sum of the drastic
and Hamming distances. Using the notation by Konieczny and
Perez~\cite{koni-pere-11}:

\begin{eqnarray*}
\Delta^{dd,[1,\ldots,1],\cdot} &=& \Delta^{d_D,\Sigma} \\
\Delta^{dh,[1,\ldots,1],\cdot} &=& \Delta^{d_h,\Sigma}
\end{eqnarray*}

The $dh$ distance was first used in belief revision by Dalal~\cite{dala-88};
for this reason, it is sometimes called ``Dalal distance''.
Revesz~\cite{reve-93,reve-97} used it with weights for belief merging, followed
by Lin and Mendelzon~\cite{lin-96,lin-mend-99}. Weights reflect the reliability
of the sources: the distance from a formula of large weight affects the total
more than the distance from a formula of low weight.

When reliability is unknown, all possible weight vectors are considered. The
set of all weight vectors is the focus of this article:

$$
W_\exists = \{[w_1,\ldots,w_m] \mid w_i \in \Z_{>0}\}
$$

Nevertheless, other sets of weight vectors are considered. In some scenarios a
source is correct and the others only provide refining information. For
example, a cardiologist, a pneumologist and an allergologist may have
contrasting opinions about the state of a patient; but if the illness is in
fact a heart disease then the cardiologist is likely to be right on everything,
for example on the reason of the breathing problems, even if that contradicts
the pneumologist and the allergologist; their opinion only provide some
additional insight. In the same way, if the problem is an allergy then the
allergologist is likely right on everything, and the same for the pneumologist.
In these cases, one source is totally correct, but which one is unknown.

$$
W_a = \{ [a,1,\ldots,1], [1,a,1,\ldots,1], \ldots, [1,\ldots,1,a] \}
$$

The value of $a$ for scenarios like that of the three doctors depends on the
maximal possible distance between a model and a formula. For the drastic
distance, $a=m+1$ suffices, where $m$ is the number of formulae to be merged.
For the Hamming distance, $a=n \times m + 1$, where $n$ is the number of
variables.

If each formula is by itself consistent with the integrity constraints, this
set of vectors satisfy the disjunction property: every model of the merged
formulae satisfy at least one of the formulae.

Finally, merging with fixed weights falls in this generalization as the set
comprising a single vector. For example, the case of equal reliability is
captured by:

$$
W_= = \{[1,\ldots,1]\}
$$

In all these cases, a set of weights $W.$ represents all possible reliability
the sources are considered to have. Three relevant such sets are $W_\exists$,
$W_a$ or $W_=$. The set $W_=$ is for equally reliable sources; $W_\exists$ is
the other extreme: the reliability of the sources is completely unknown. Every
$W \in W.$ is an encoding of the reliability of the sources. All of these are
plausible alternatives. Every scenario (every model) that is possible when
merging with some $W \in W.$ is possible when merging with $W.$:

$$
\Delta^{d,W.,\cdot}_\mu(F_1,\ldots,F_m)
=
\bigcup_{W \in W.} \Delta^{d,W,\cdot}_\mu(F_1,\ldots,F_m)
$$

This definition is parameterized by a distance $d$ (which could be $dd$ or
$dh$), a set of weight vectors $W.$ (usually $W_\exists$ but may also be $W_a$
or $W_=$) and an aggregation function (always the scalar product $\cdot$ in this
article). Given some formulae $F_1,\ldots,F_m$, this is how they are merged
under integrity constraints $\mu$.


\section{Dominance}

If a model is further from every formula than another, the latter is always
preferred to the former regardless of the weights. The second model {\em
dominates} the first. Despite the seeming triviality of the concept, a number
of relevant results follow:

\begin{itemize}

\item if a model has minimal weighted distance for some weights, it is not
strictly dominated by another;

\item if the codomain is binary, the converse also holds: a model that is not
strictly dominated by another has minimal weighted distance for some weights;

\item for a ternary codomain, there exists two formulae such that their merge
does not include an undominated model;

\item there is a corner case of a ternary codomain with two formulae where
undominance still implies minimality.

\end{itemize}

Rather than defining dominance over models with respect to formulae, the
condition has a simpler formalization over vectors of integers. It can then be
carried over to the distance vectors of two models.

\begin{definition}
\label{dominance}

A vector of integers $D$ dominates another $D'$, denoted $D \leq D'$, if every
element of $D$ is less than or equal than the element of the same index in
$D'$. Strict dominance is the strict part of this ordering: $D < D'$ if $D \leq
D'$ and $D' \not\leq D'$.

\end{definition}

If the distance vector of a model is strictly dominated by that of another, the
first is never minimal regardless of the weights. This fact holds because
weights are (strictly) positive.

\begin{lemma}
\label{dominance-always}

For every distance $d$, vector of weights $W \in W_\exists$ and model $I$,
{} if $I \in \Delta^{d,W,\cdot}_\mu(F_1,\ldots,F_m)$
{} then $d(J,F_1,\ldots,F_m) < d(I,F_1,\ldots,F_m)$
{} holds for no model $J$ of $\mu$.

\end{lemma}

\proof The claim is proved in the opposite direction:
{} $d(J,F_1,\ldots,F_m) < d(I,F_1,\ldots,F_m)$
entails
{} $I \not\in \Delta^{d,W,\cdot}_\mu(F_1,\ldots,F_m)$.

Since weights are all strictly positive,
{} $d(J,F_1,\ldots,F_m) \leq d(I,F_1,\ldots,F_m)$
entails
{} $W \cdot d(J,F_1,\ldots,F_m) \leq W \cdot d(I,F_1,\ldots,F_m)$
and
{} $d(I,F_1,\ldots,F_m) \not\leq d(J,F_1,\ldots,F_m)$
entails
{} $W \cdot d(I,F_1,\ldots,F_m) \not\leq W \cdot d(J,F_1,\ldots,F_m)$.
These two consequences together are
{} $W \cdot d(J,F_1,\ldots,F_m) < W \cdot d(I,F_1,\ldots,F_m)$,
which proves that $I$ is not a model of minimal distance weighted by $W$, and
is not therefore in $\Delta^{d,W,\cdot}_\mu(F_1,\ldots,F_m)$.~\qed

This result is almost trivial, since merging select models that have a minimal
value of the sum of the distances, each multiplied by a positive weight. The
converse does not hold in general, but does in a relevant case: when 
$d(I,F_i)$ can only be $0$ or $1$, or more generally when the codomain of $d$
has size two.

\begin{lemma}
\label{dominance-only}

{} If the codomain of $d(.,.)$ is a subset of cardinality two of $\Z_{\geq 0}$,
{} $I$ is a model of $\mu$ and
{} $d(I,F_1,\ldots,F_m)$ is not strictly dominated
by the vector of distances of another model of $\mu$,
{} then there exists $W$
{} such that $I \in \Delta^{d,W,\cdot}_\mu(F_1,\ldots,F_m)$.

\end{lemma}

\proof Let $\{a,b\}$ be the codomain of $d$, where $a < b$ without loss of
generality. The $i$-th weight of $W$ is $w_i = m+1$ if $d(I,F_i) = a$, and $w_i
= 1$ otherwise.

For every other model $J$ of $\mu$, the weighted distance of $J$ is proved to
be greater than or equal to that of $I$. Two cases are possible: either
$d(J,F_i) = a$ for every $F_i$ such that $d(I,F_i) = a$, or this is not the
case for at least one formula $F_i$.

In the first case, $d(J,F_i) \leq d(I,F_i)$ for every $F_i$, which implies
{} $d(J,F_1,\ldots,F_m) \leq d(I,F_1,\ldots,F_m)$.
Since $J$ does not dominate $I$ by assumption,
{} $d(I,F_1,\ldots,F_m) \not\leq d(J,F_1,\ldots,F_m)$
is false, which means that
{} $d(I,F_1,\ldots,F_m) \leq d(J,F_1,\ldots,F_m)$ 
is true. The distance vectors of $I$ and $J$ are the same. Therefore,
multiplying each by $W$ produces the same result. This proves that $I$ is
minimal.

The second case is that $d(I,F_i) = a$ and $d(J,F_i) = b$ for some $F_i$. If
$k$ is the number of formulae $F_i$ such that $d(I,F_i) = a$, the weighted
distance of $I$ is

\begin{eqnarray*}
W \cdot d(I,F_1,\ldots,F_m) &=&
(m+1) \times k \times a + 1 \times (m-k) \times b		\\
&=&
m \times k \times a + 1 \times k \times a +			
1 \times m \times b - 1 \times k \times b			\\
&=& 
m \times k \times a + k \times a +
m \times b - k \times b						\\
&<&
m \times k \times a + m \times b
\end{eqnarray*}

Only $d(J,F_i) = b$ is known, the distance of $J$ from the other formulae may
be either $a$ or $b$. Assuming it is $a$ for all of them leads to the minimal
possible weighted distance, which is:

\begin{itemize}

\item one formula has distance $b$, but since $d(I,F_i) = a$ the weight is
$m+1$;

\item the other formulae have all distance $a$, but

\begin{itemize}

\item the $k-1$ formulae for which $d(I,F_i) = a$ have weight $m+1$;

\item the remaining $m-k$ formulae have weight $1$.

\end{itemize}

\end{itemize}

The weighted distance of $J$ is therefore:

\begin{eqnarray*}
W \cdot d(J,F_1,\ldots,F_m)
&\geq&
(m+1) \times 1 \times b +
(m+1) \times (k-1) \times a +
1 \times (m-k) \times a
\\
&=&
m \times 1 \times b + 1 \times 1 \times b +
m \times k \times a - m \times 1 \times a + 1 \times k \times a - a +
\\
&&
1 \times m \times a - 1 \times k \times a
\\
&=&
m \times b + b +
m \times k \times a - m \times a + k \times a - a +
m \times a - k \times a
\\
&=&
m \times b + b + m \times k \times a - a
\\
&>&
m \times b + m \times k \times a
\end{eqnarray*}

The weighted distance of $I$ is less than this, as shown above.~\qed

The two last lemmas imply that the minimal models according to a distance of
binary codomain are exactly the models whose distance vector is not dominated
by another.

\begin{theorem}
\label{binary}

If the codomain of $d(.,.)$ is a subset of cardinality two of $\Z_{\geq 0}$,
then $\Delta^{d,W_\exists,\cdot}_\mu(F_1,\ldots,F_m)$ is the set of all models
of $\mu$ of minimal distance vector according to the dominance ordering.

\end{theorem}

\proof Lemma~\ref{dominance-always} proves that models of minimal weighted
distance are never strictly dominated by any other model of $\mu$. By
Lemma~\ref{dominance-only}, if the codomain of $d$ is binary then every model
that is not strictly dominated has a weight vector $W$ that makes its weighted
distance minimal. Since $W_\exists$ contains all weight vectors, the claim is
proved.~\qed

The next question is whether this condition holds for every fixed-size
codomain, or whether a codomain of size three is sufficient for making some
undominated model to be excluded from merge. The latter is indeed the case in
general. A preliminary lemma will be useful in the sequel.

\begin{lemma}
\label{undominated-excluded}

If $\mu$ has three models of distance $[3,0]$, $[2,2]$ and $[0,3]$ from $F_1$
and $F_2$, then $\Delta^{dh,W_\exists,\cdot}_\mu(F_1,F_2)$ does not contain the
model at distance $[2,2]$.

\end{lemma}

\proof If the model were minimal, the following set of linear inequalities
would be satisfiable for some $W=[w_1,w_2]$.

\begin{eqnarray*}
w_1 \times 2 + w_2 \times 2 &\leq& w_1 \times 3 + w_2 \times 0 \\
w_1 \times 2 + w_2 \times 2 &\leq& w_1 \times 0 + w_2 \times 3
\end{eqnarray*}

The first implies $w_2 \times 2 \leq w_1$, the second $w_1 \times 2 \leq w_2$:
each weight is at least twice the other. No positive values may satisfy
both.~\qed

This is almost the proof of the claim, but the codomain of the Hamming distance
has size unbounded, not three. However, a slight change in the definition of
the distance and a comparison of the distance vector are enough.

\begin{theorem}

For some distance $d$ with codomain of size three, there exists $I$, $\mu$ and
$F_1,\ldots,F_m$ such that $I \models \mu$ and
{} $I \not\in \Delta^{d,W_\exists,\cdot}_\mu(F_1,\ldots,F_m)$ but
{} $d(J,F_1,\ldots,F_m) < d(I,F_1,\ldots,F_m)$ does not hold
for any $J \models \mu$.

\end{theorem}

\proof This is shown on the codomain $\{0,2,3\}$ and a formula $\mu$ with three
models of distance vectors $[0,3]$, $[2,2]$ e $[3,0]$ from $F_1$ and $F_2$.
That such formulae exist is later proved by Lemma~\ref{distances-formulae} for
the Hamming distance. To obtain the right codomain $\{0,2,3\}$ the distance is
modified by setting $dh'(I,F_i)=3$ for every model $K$ such that $dh(K,F_i)
\not\in \{0,2,3\}$. This change does not affect the distance of the model under
consideration.

None of the three distance vectors is strictly dominated by another. However,
the previous lemma shows that $[2,2]$ is not minimal for any weight
vector.~\qed

The proof of this counterexample is based on a function of codomain
$\{0,2,3\}$, which does not look very natural, as $\{0,1,2\}$ would. But the
counterexample does not work in this case, at least for two formulae. The proof
of this claim is not much long but tedious, and is omitted. The counterexample
holds again with the same codomain but three formulae.

\section{Drastic distance}

Merging with all possible weights and the drastic distance is the same as
disjoining all maximal subsets of $F_1,\ldots,F_m$ that are consistent with
$\mu$. This is proved in three steps:

\begin{itemize}

\item dominance with the drastic distance is the same as the containment of the
set of formulae $F_1,\ldots,F_m$ satisfied by the models;

\item the models of the maxcons are the models that are minimal according to
that containment;

\item therefore, the models of the maxcons are exactly the undominated models;
by the results in the previous section, they are the models of minimal weighted
distance according to some weights.

\end{itemize}

Maximal consistent subsets (maxcons) have a general definition over lists of
sets of formulae, but what is necessary for this article is only the version
with a list of two sets, the first comprising a single consistent formula $\mu$
and the second $F_1,\ldots,F_m$. With this limitation, the (possibly
non-maximal) consistent subset and the maximal consistent subsets are defined
as:

\begin{eqnarray*}
\con_\mu(F_1,\ldots,F_m) &=& \{
	S \subseteq \{\mu,F_1,\ldots,F_m\}
	\mid
	\mu \in S
	\mbox{ and }
	S \not\models \bot
\} \\
\maxcon_\mu(F_1,\ldots,F_m) &=& \{
	S \in \con_\mu(F_1,\ldots,F_m)
	\mid
	\not\exists S' \in \con_\mu(F_1,\ldots,F_m) ~.~
		S \subset S'
\}
\end{eqnarray*}

Since $\mu$ is consistent, these sets cannot be empty. To establish the
correspondence between models and maxcons, the subset of formulae satisfied by
a model is needed.

\begin{definition}

The set of formulae satisfied by a model $I$ is denoted
$\subsat(I,F_1,\ldots,F_m) = \{ F_i \mid I \models F_i\}$.

\end{definition}

The basic brick in the proof construction is that dominance of the drastic
distance vectors is the same as containment of the subsets of formulae
satisfied by models.

\begin{lemma}
\label{subsat-drastic}

For every pair of models $I$ and $J$ and every formulae $F_1,\ldots,F_m$, the
following two conditions are equivalent, where $dd$ is the drastic distance:

\begin{eqnarray*}
dd(I,F_1,\ldots,F_m)		&\leq&		dd(J,F_1,\ldots,F_m)	\\
\subsat(J,F_1,\ldots,F_m)	&\subseteq&	\subsat(I,F_1,\ldots,F_m)
\end{eqnarray*}

\end{lemma}

\proof The claim is first proved on a single formula $F_i$, and then extended
to multiple formulae. In the case of a single formula, the claim is that
{} $dd(I,F_i) \leq dd(J,F_i)$
is the same as
{} $\subsat(J,F_i) \subseteq \subsat(I,F_i)$

Both $dd()$ and $\subsat()$ are defined in terms of $I \models F_i$ and $J
\models F_i$. Each of these two conditions can be true or false, depending on
the models $I$ and $J$ and the formula $F_i$. The four cases are:

$$
\begin{array}{cc|cccc}
&& dd(I,F_i) & dd(J,F_i) & \subsat(I,F_i) & \subsat(I,F_i)		\\
\hline
I \models F_i & J \models F_i & 0 & 0 & \{F_i\} & \{F_i\}		\\
I \models F_i & J \not\models F_i & 0 & 1 & \{F_i\} & \emptyset		\\
I \not\models F_i & J \models F_i & 1 & 0 & \emptyset & \{F_i\}		\\
I \not\models F_i & J \not\models F_i & 1 & 1 & \emptyset & \emptyset
\end{array}
$$

Only in the third cases $dd(I,F_i) \leq dd(J,F_i)$ is false, and this is also
the only case where $\subsat(J,F_i) \subseteq \subsat(I,F_i)$ is false. This
proves the claim for a single formula.

\

By definition, $dd(I,F_1,\ldots,F_m) \leq dd(J,F_1,\ldots,F_m)$ is the same as
$dd(I,F_i) \leq dd(J,F_i)$ for every $F_i$. The claim is proved if
{} $\subsat(J,F_1,\ldots,F_m) \subseteq \subsat(I,F_1,\ldots,F_m)$
is the same as $\subsat(J,F_i) \subseteq \subsat(I,F_i)$ for every $F_i$.

By definition,
{} $\subsat(I,F_1,\ldots,F_m) = \bigcup_{1 \leq i \leq m} \subsat(I,F_i)$,
and the same for $J$. If $\subsat(J,F_i) \subseteq \subsat(I,F_i)$ for every
$F_i$, then $\subsat(J,F_1,\ldots,F_m) \subseteq \subsat(I,F_1,\ldots,F_m)$
holds.

The converse holds because each $\subsat(J,F_i)$ and $\subsat(I,F_i)$ may only
contain $F_i$. As a result, $\subsat(J,F_i) \not\subseteq \subsat(I,F_i)$ only
holds if $\subsat(J,F_i) = \{F_i\}$ and $\subsat(I,F_i) = \emptyset$. Since
$\subsat(I,F_i)$ is the only part of $\subsat(I,F_1,\ldots,F_m)$ that may
contain $F_i$, this set does not contain $F_i$. Instead,
$\subsat(J,F_1,\ldots,F_m)$ contains $F_i$ because this formula is in
$\subsat(J,F_i)$. This proves that if $\subsat(J,F_i) \not\subseteq
\subsat(I,F_i)$ for some $F_i$ then
{} $\subsat(J,F_1,\ldots,F_m) \not\subseteq \subsat(I,F_1,\ldots,F_m)$.

Overall, $\subsat(J,F_1,\ldots,F_m) \subseteq \subsat(I,F_1,\ldots,F_m)$ is
equivalent to $\subsat(J,F_i) \subseteq \subsat(I,F_i)$ for every $i$. These
conditions have been previously proved to be equivalent to $dd(I,F_i) \leq
dd(J,F_i)$ each. Together, these define $dd(I,F_1,\ldots,F_m) \leq
dd(J,F_1,\ldots,F_m)$.~\qed

This lemma links the dominance ordering under $dd$ and the containment of
$\subsat$. What is left is a lemma that links the latter with the maxcons.

\begin{lemma}
\label{subsat-maxcon}

A model $I$ of $\mu$ satisfies some element of $\maxcon_\mu(F_1,\ldots,F_m)$ if
and only if $\subsat(I,F_1,\ldots,F_m) \subset \subsat(J,F_1,\ldots,F_m)$ holds
for no model $J$ of $\mu$.

\end{lemma}

\proof Let $I$ be a model of $\mu$ such that $\subsat(I,F_1,\ldots,F_m) \subset
\subsat(J,F_1,\ldots,F_m)$ holds for no model $J$ of $\mu$. The set
$\subsat(I,F_1,\ldots,F_m)$ will be proved to be a maxcon. This set is
consistent with $\mu$ because $I$ satisfies both. It is also maximally so.
Otherwise,
{} $\subsat(I,F_1,\ldots,F_m) \cup \{\mu,F_i\}$
would be consistent for some
{} $F_i \not\in \subsat(I,F_1,\ldots,F_m)$.
Consistency implies the existence of a model $J \models
\subsat(I,F_1,\ldots,F_m) \cup \{\mu,F_i\}$. Since $J$ satisfies all these
formulae, $\subsat(J,F_1,\ldots,F_m)$ contains all of them:
{} $\subsat(I,F_1,\ldots,F_m) \cup \{F_i\} \subseteq
{}  \subsat(J,F_1,\ldots,F_m)$.
This implies $\subsat(I,F_1,\ldots,F_m) \subset \subsat(J,F_1,\ldots,F_m)$ for
a model $J$ that also satisfies $\mu$, contrary to assumption.

Let $J$ be a model of $\mu$ such that $\subsat(I,F_1,\ldots,F_m) \subset
\subsat(J,F_1,\ldots,F_m)$. The claim is that $I$ is not in any maxcon. By
contradiction, let $M$ be such a maxcon. Since all its formulae satisfy $I$, it
holds $M \subseteq \subsat(I,F_1,\ldots,F_m)$. By assumption, this set is
strictly contained in $\subsat(J,F_1,\ldots,F_m)$ for some $J \models \mu$.
Since $J$ satisfies both $\subsat(J,F_1,\ldots,F_m)$ and $\mu$, this other set
$\subsat(J,F_1,\ldots,F_m)$ is consistent with $\mu$, contradicting the
assumption that $M$ is maximally consistent with $\mu$.~\qed

The lemma is the final piece for the construction of the link between maxcons
and dominance under the drastic distance.

\begin{theorem}
\label{drastic-maxcons}

For every consistent formulae $\mu, F_1,\ldots,F_m$,

$$
\Delta^{dd,W_\exists,\cdot}_\mu(F_1,\ldots,F_m) =
\bigvee \maxcon_\mu(F_1,\ldots,F_m)
$$

\end{theorem}

\proof By definition, all elements of $\maxcon_\mu(F_1,\ldots,F_m)$ contain
$\mu$. Therefore, the models of
{} $\bigvee \maxcon_\mu(F_1,\ldots,F_m)$
all satisfy $\mu$, making Lemma~\ref{subsat-maxcon} applicable:
{} $\bigvee \maxcon_\mu(F_1,\ldots,F_m)$
are exactly the models $I$ of $\mu$ such that
{} $\subsat(J,F_1,\ldots,F_m) \subset \subsat(I,F_1,\ldots,F_m)$
holds for no other model $J$ of $\mu$. This is equivalent to
$dd(J,F_1,\ldots,F_m) < dd(I,F_1,\ldots,F_m)$ by Lemma~\ref{subsat-drastic}.
Therefore, these models $I$\plural are the models of $\mu$ that are not
strictly dominated by other models of $\mu$. Since the codomain of $dd$ is
binary, these are the models of
{} $\Delta^{dd,W_\exists,\cdot}_\mu(F_1,\ldots,F_m)$
by Lemma~\ref{binary}.~\qed

Maxcons have been long used in belief revision~%
\cite{resc-mano-70,brew-89,gins-86,fagi-ullm-vard-83,bara-etal-92,%
benf-etal-97,koni-pere-11,ammo-etal-15,gran-hunt-11,dubo-etal-16}. Yet, they
are sometimes dismissed as ``unsuitable for merging'' because they do not take
into account the distribution of information among the
sources~\cite{koni-00,koni-pere-11}. This theorem redeems them: maxcons are
weighted merge with the drastic distance and unknown reliability. Not only they
are suitable for merging, they deal with the common situation where the
credibility of the sources cannot be assessed.

An example clarifies why. If reliability is unknown, a formula $\neg x$
provided by two sources cannot beat a formula $x$ provided by one source, since
the one source may be much more reliable than the others as far as it is known.
Merging by maxcons collects as many formulae as possible while retaining
consistency; each maxcon may come from the most reliable sources, making the
number of formulae itself irrelevant.

\section{Hamming distance}

The Hamming distance $dh$ has a codomain of more than two elements. Therefore,
the previous results about binary codomains do not apply. Some existence
results are proved:

\begin{itemize}

\item every given set of distance vectors is obtainable from some formulae
$\mu,F_1,\ldots,F_m$;

\item there are $\mu,F_1,\ldots,F_m$ such that merging with all possible weight
vectors is only equivalent to merging with some sets of weights of at least
exponential cardinality.

\end{itemize}

Merging with the Hamming distance does not have a simple equivalent form like
for the drastic distance, which selects the models that are not strictly
dominated by others. The same does not hold in general: a model that is
undominated may still be excluded in the merging. This was proved abstractly by
three distance vectors $[3,0]$, $[2,2]$ e $[0,3]$. The Hamming distance allows
proving that such distance vectors can be obtained from concrete formulae. More
generally, every set of distance vectors can. This existence result is
analogous to a similar one for maxcons~\cite[Lemma~4.6]{libe-15}.

\begin{lemma}
\label{distances-formulae}

Given some vectors of distances $D_1,\ldots,D_o$ of $m$ elements each, all
bounded by an integer $n$, for some formulae $\mu$ and $F_1,\ldots,F_m$ over $n
\times m$ variables the vectors of distance from the models of $\mu$ to
$F_1,\ldots,F_m$ are exactly $D_1,\ldots,D_o$.

\end{lemma}

\proof Formulae $\mu$ and $F_1,\ldots,F_m$ are build over the set of variables
{} $\{x_j^i \mid 1 \leq j \leq n ,~ 1 \leq i \leq m\}$.
Each formula $F_i$ is a conjunction of some of them.

$$
F_i = x_1^i \wedge \cdots \wedge x_n^i
$$

Given a model $I$, its closest model of $F_i$ has all variables
$x_1^i,\ldots,x_n^i$ positive and the same evaluation of $I$ on the other
variables. Therefore, $dh(I,F_i)$ is the number of variables
$x_1^i,\ldots,x_n^i$ assigned false by $I$.

For each distance vector $[d_1,\ldots,d_m]$ among the given ones, $\mu$ has the
following model:

$$
\bigcup_{1 \leq i \leq m}
	\{\neg x_j^i \mid 1 \leq j \leq d_i\} \cup
	\{x_j^i \mid d_i < j \leq n\}
$$

For each $i$, this model has $d_i$ negative variables among
$x_1^i,\ldots,x_n^i$; therefore, $dh(I,F_i) = d_i$. As a result,
$dh(I,F_1,\ldots,F_m) = [d_1,\ldots,d_m]$. Since $\mu$ has one such model for
each of the given distance vectors, the claim is proved.~\qed

This theorem allows for an easy way for building counterexamples: rather than
providing $d$ and $\mu,F_1,\ldots,F_m$ that have a certain property, one may
simply show that the same property holds on a set of distance vectors. This
method was already used to prove that some undominated models are not selected
by merging, for some distance. In particular, it shows this being the case for
the Hamming distance.

Another application is the proof that an exponential number of weight vectors
have to be considered when merging. The definition itself requires all weight
vectors to be taken into account: a model is selected if and only if it is
selected by at least one of the infinitely many weight vectors in $W_\exists$.
The next lemma shows that at least exponentially many have to be considered. It
will be later proved that an exponential number suffices.

\begin{lemma}
\label{two}

There exists three formulae $\mu,F,F'$ on an alphabet of six variables such
that every $W_r$ such that
{} $\Delta^{dh,W_r,\cdot}(F,F') = \Delta^{dh,W_\exists,\cdot}(F,F')$
contains at least two weight vectors.

\end{lemma}

\proof By Lemma~\ref{distances-formulae}, given distances $[3,0]$, $[1,1]$ and
$[0,3]$, there exists formulae $\mu,F_1,F_2$ over six variables such that the
three models of $\mu$ have these distance vectors.

All three distance vectors are minimal for some $W \in W_\exists$. In
particular, the first two are minimal for $W=[2,4]$, the third is minimal for
$W=[4,1]$. This proves that $\Delta^{dh,W_\exists,\cdot}(F_1,\ldots,F_m)$
contains all three models of $\mu$.

Contrary to the claim, a single weight vector is assumed to produce the same
result. Since the model at distance $[3,0]$ is minimal, its weighted distance
is less than or equal to that of the model at distance $[1,1]$, and the same
for $[0,3]$:

\begin{eqnarray*}
\relax	[w_1,w_2] \cdot [3,0]	&\leq&	[w_1,w_2] \cdot [1,1]	\\
\relax	[w_1,w_2] \cdot [0,3]	&\leq&	[w_1,w_2] \cdot [1,1]
\end{eqnarray*}

Expressing the two vector products explicitly:

\begin{eqnarray*}
w_1 \times 3	&\leq&	w_1 + w_2	\\
w_2 \times 3	&\leq&	w_1 + w_2
\end{eqnarray*}

Since all weights are positive, the left-hand and right-hand sides of these
inequalities can be added, leading to
{} $w_1 \times 3 + w_2 \times 3 \leq w_1 \times 2 + w_2 \times 2$,
which is impossible for positive weights. This proves that no single weight
vector produces the same merging of $W_\exists$.~\qed

This lemma shows a pair of formulae that can be merged using two weight vectors
rather than infinitely many. Still better, it shows that at least two weight
vectors are necessary. The construction can be replicated over many distinct
alphabets of six variables each. Each alphabet doubles the number of necessary
weight vectors, leading to exponentiality.

\begin{lemma}
\label{exponential-minimal}

There exists $\mu,F_1,\ldots,F_m$ such that the size of every $W_r$ for which
{} $\Delta^{dh,W_r,\cdot}(F_1,\ldots,F_m) =
{}  \Delta^{dh,W_\exists,\cdot}(F_1,\ldots,F_m)$
is exponential in the size of the formulae.

\end{lemma}

\proof By Lemma~\ref{two}, there exists formulae $\mu,F,F'$ on six variables
$X$ such that $W_\exists$ is only equivalent to sets weight vectors of
cardinality greater than or equal to two. Since the variables are six, these
three formulae are equivalent to formulae of size at most $2^6$, a constant.

This construction can be replicated on $\frac{m}{2}$ disjoint alphabets
$X_1,\ldots,X_\frac{m}{2}$ of six variables each, leading to $\frac{m}{2}$
triples $\mu_i,F_i,F_i'$ of formulae with no shared variables among different
triples. The size of each triple is bounded by a constant.

Each $\mu_i$ has three models. Therefore, $\bigwedge \mu_i$ has $3^\frac{m}{2}$
models. Each of these models has a distance vector that is a combination of
$\frac{m}{2}$ subvectors, each being $[3,0]$, $[1,1]$ or $[0,3]$. For each
part, either of two weight vectors are needed to make the model minimal. The
total number of necessary weight vectors is therefore exponential in $m$.~\qed

This result relies on an unbounded number of formulae to be merged. With two
formulae, a number of weight vectors linear in the number of variables
suffices. Such a result is straightforward when seen on the graphical
representation of merge as shown in the next section.

\section{Finiteness}

Merging with unknown reliability involves the infinite set of weight vectors
$W_\exists$. However, only an exponential number of vectors matter.
Furthermore, minimality of a model can be recast as a comparison with a bounded
number of other models. In summary:

\begin{itemize}

\item given $\mu,F_1,\ldots,F_m$, merging with a (possibly infinite) set of
weight vectors is the same as merging with at most exponentially many weight
vectors;

\item a graphical representation of merging two formulae clarifies the
difference between dominance, minimality under fixed weights and all possible
weights;

\item minimality of the distance from a model of $\mu$ to $F_1,\ldots,F_m$ for
some weight vectors can be checked by comparing the model with $m$ other models
of $\mu$ at time;

\item merging $\mu,F_1,\ldots,F_m$ with all possible weight vectors can be
expressed in terms of merging $\mu',F_1,\ldots,F_m$ for all formulae $\mu'$
that are satisfied by at most $m+1$ models of $\mu$.

\end{itemize}

When merging $F_1,\ldots,F_m$ with constraints $\mu$, a set of weight vectors
$W.$ may be equivalent to another $W_r$ if they produce the same result. This
is specific to the formulae to be merged and integrity constraints: the same
sets may not be equivalent when changing $\mu$ or any of the formulae $F_i$.
This is still interesting because the infinite set of vectors $W_\exists$ can
be shown to always be the same as an exponentially sized set.

\begin{lemma}

For every $\mu$, $F_1,\ldots,F_m$ and set of weight vectors $W.$ there exists a
set of weight vectors $W_r$ of size bounded by an exponential in the size of
the formulae such that:

$$
\Delta^{d,W_r,\cdot}(F_1,\ldots,F_m)
=
\Delta^{d,W.,\cdot}(F_1,\ldots,F_m)
$$

\end{lemma}

\proof By definition,
{} $I \in \Delta^{d,W.,\cdot}(F_1,\ldots,F_m)$
if $W.$ contains a vector $W$ such that
{} $I \in \Delta^{d,W,\cdot}(F_1,\ldots,F_m)$.
In the same way,
{} $I \in \Delta^{d,W_r,\cdot}(F_1,\ldots,F_m)$
if $W_r$ contains a vector $W$ with the same property.
If $W_r \subseteq W.$, every model of $\Delta^{d,W_r,\cdot}(F_1,\ldots,F_m)$
is also a model of $I \in \Delta^{d,W.,\cdot}(F_1,\ldots,F_m)$. The converse
can be proved for a suitable choice of $W_r \subseteq W.$.

No matter what $W.$ contains, $\Delta^{d,W.,\cdot}(F_1,\ldots,F_m)$ is always a
set of models on the alphabet of the formulae $\mu,F_1,\ldots,F_m$, and the
models on this alphabet are exponentially many.

By definition, $I \in \Delta^{d,W.,\cdot}(F_1,\ldots,F_m)$ if there exists at
least one vector $W \in W.$ such that
{} $I \in \Delta^{d,W,\cdot}(F_1,\ldots,F_m)$.
Let $\preceq$ be the lexicographic order. The set $W_r$ is:

$$
W_r =
\bigcup_{I \in \Delta^{d,W.,\cdot}(F_1,\ldots,F_m)}
	\min_\preceq \{ 
		W
		\mid
		I \in \Delta^{d,W,\cdot}(F_1,\ldots,F_m),
	\}
$$

By construction, for every model $I \in \Delta^{d,W.,\cdot}(F_1,\ldots,F_m)$
this set $W_r$ contains at least a vector $W$ such that $I \in
\Delta^{d,W,\cdot}(F_1,\ldots,F_m)$. As a result, $I$ is also in $I \in
\Delta^{d,W_r,\cdot}(F_1,\ldots,F_m)$.~\qed

Lemma~\ref{exponential-minimal} shows a situation where the necessary number of
weight vectors to consider is at least exponential. It is in a way the converse
of the last theorem: no more than exponentially many vectors are required, and
in some cases that many are required.

\

An enlightening graphical representation of the distance vectors helps to
understand which models are selected by merging. Only two formulae $F_1$ and
$F_2$ are considered along with constraints $\mu$. Each model is drawn as a
point at coordinates $[d(I,F_1),f(I,F_2)]$.
As a result, the models of $\mu$ are points in the first quadrant of the
Cartesian plane. Theorem~\ref{distances-formulae} tells that every set of
points represent the models of some formula $\mu$, their coordinates being the
Hamming distances from some formulae $F_1$ and $F_2$.

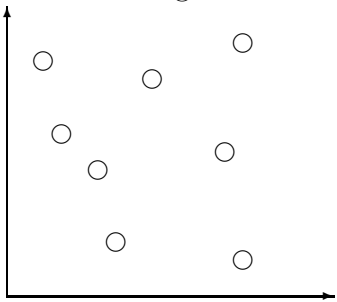
\setlength{\unitlength}{5000sp}%
\begingroup\makeatletter\ifx\SetFigFont\undefined%
\gdef\SetFigFont#1#2#3#4#5{%
  \reset@font\fontsize{#1}{#2pt}%
  \fontfamily{#3}\fontseries{#4}\fontshape{#5}%
  \selectfont}%
\fi\endgroup%
\begin{picture}(1674,1554)(4639,-4123)
\thinlines
{\color[rgb]{0,0,0}\put(4681,-4021){\vector( 0, 1){1440}}
\put(4681,-4021){\vector( 1, 0){1620}}
}%
{\color[rgb]{0,0,0}\put(5131,-3391){\circle{90}}
}%
{\color[rgb]{0,0,0}\put(5221,-3751){\circle{90}}
}%
{\color[rgb]{0,0,0}\put(4951,-3211){\circle{90}}
}%
{\color[rgb]{0,0,0}\put(4861,-2851){\circle{90}}
}%
{\color[rgb]{0,0,0}\put(5401,-2941){\circle{90}}
}%
{\color[rgb]{0,0,0}\put(5851,-2761){\circle{90}}
}%
{\color[rgb]{0,0,0}\put(5761,-3301){\circle{90}}
}%
{\color[rgb]{0,0,0}\put(5851,-3841){\circle{90}}
}%
\end{picture}%
\nop{
|
|     o
|           o
|       o
|            o
|            
|             o
|                                       .
|                   
|                                       .
|                     o
|
+----------------------------
}

The origin is the ideal position since there the distance from both formulae is
zero. If a model is in the origin, it is always minimal. Otherwise, models are
preferred when close to the origin.

A model is undominated if it and the origin are the opposite corners of a
rectangle free of other models. This is a simple but crude condition that
ensures that a model is close to the origin. Still, other models may be
geometrically closer than it is.

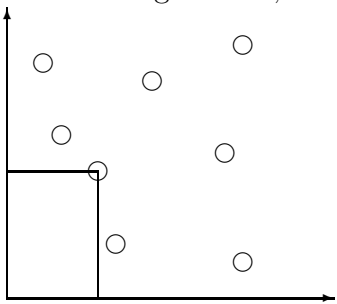
\setlength{\unitlength}{5000sp}%
\begingroup\makeatletter\ifx\SetFigFont\undefined%
\gdef\SetFigFont#1#2#3#4#5{%
  \reset@font\fontsize{#1}{#2pt}%
  \fontfamily{#3}\fontseries{#4}\fontshape{#5}%
  \selectfont}%
\fi\endgroup%
\begin{picture}(1674,1554)(4639,-4123)
\thinlines
{\color[rgb]{0,0,0}\put(4681,-4021){\vector( 0, 1){1440}}
\put(4681,-4021){\vector( 1, 0){1620}}
}%
{\color[rgb]{0,0,0}\put(5131,-3391){\circle{90}}
}%
{\color[rgb]{0,0,0}\put(5221,-3751){\circle{90}}
}%
{\color[rgb]{0,0,0}\put(4951,-3211){\circle{90}}
}%
{\color[rgb]{0,0,0}\put(4861,-2851){\circle{90}}
}%
{\color[rgb]{0,0,0}\put(5401,-2941){\circle{90}}
}%
{\color[rgb]{0,0,0}\put(5851,-2761){\circle{90}}
}%
{\color[rgb]{0,0,0}\put(5761,-3301){\circle{90}}
}%
{\color[rgb]{0,0,0}\put(5851,-3841){\circle{90}}
}%
{\color[rgb]{0,0,0}\put(4681,-3391){\line( 1, 0){450}}
\put(5131,-3391){\line( 0,-1){630}}
}%
\end{picture}%
\nop{
|
|      o
|           o
|       o
|------------o
|            |
|            |o
|            |                          .
|            |      
|            |                          .
|            |        o
|            |
+----------------------------
}

Given an arbitrary integer $a$, a model has weighted distance $a$ if
{} $w_1 \times d(I,F_1) + w_2 \times d(I,F_2) = a$.
Since $d(I,F_1)$ and $d(I,F_2)$ are coordinates in the plane, this is the
equation of a line in the plane characterized by the three parameters
$w_1,w_2,a$. In other words, a model has weighted distance $a$ if and only if
is on the line. Since the three numbers $w_1,w_2,a$\plural are all positive,
the line has negative slope.

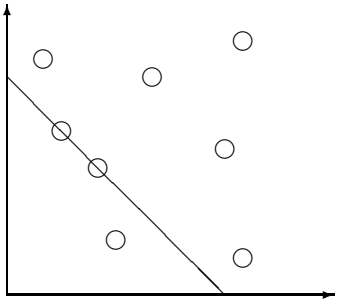
\setlength{\unitlength}{5000sp}%
\begingroup\makeatletter\ifx\SetFigFont\undefined%
\gdef\SetFigFont#1#2#3#4#5{%
  \reset@font\fontsize{#1}{#2pt}%
  \fontfamily{#3}\fontseries{#4}\fontshape{#5}%
  \selectfont}%
\fi\endgroup%
\begin{picture}(1674,1554)(4639,-4123)
\thinlines
{\color[rgb]{0,0,0}\put(4681,-4021){\vector( 0, 1){1440}}
\put(4681,-4021){\vector( 1, 0){1620}}
}%
{\color[rgb]{0,0,0}\put(5131,-3391){\circle{90}}
}%
{\color[rgb]{0,0,0}\put(5221,-3751){\circle{90}}
}%
{\color[rgb]{0,0,0}\put(4951,-3211){\circle{90}}
}%
{\color[rgb]{0,0,0}\put(4861,-2851){\circle{90}}
}%
{\color[rgb]{0,0,0}\put(5401,-2941){\circle{90}}
}%
{\color[rgb]{0,0,0}\put(5851,-2761){\circle{90}}
}%
{\color[rgb]{0,0,0}\put(5761,-3301){\circle{90}}
}%
{\color[rgb]{0,0,0}\put(5851,-3841){\circle{90}}
}%
{\color[rgb]{0,0,0}\put(4681,-2941){\line( 1,-1){1080}}
}%
\end{picture}%
\nop{
|    \                                  .
|      o                                .
|        \  o                           .
|       o  \                            .
|            o                          .
|              \                        .
|             o  \                      .
|                  \                    .
|                    \                  .
|                      \                .
|                     o  \              .
|                          \            .
+----------------------------
}

Every descending line that intersects $[d(I,F_1),d(I,F_2)]$ is a set of models
having the same weighted distance as $I$ according to some weight vector
$[w_1,w_2]$. Such a line divides the plane into two half-planes; the one where
the origin is comprises exactly the models that have a lower weighted distance
than $I$. In the figure, three points are in this half-plane. This means that
the two points the line intersects are not minimal according to the weights
$[w_1,w_2]$.

Merging by equal weights finds the line at $45$ degrees that crosses some
points but does not leave any other in the half-plane where the origin lies.
With fixed but differing weights the slope of the line changes but the
mechanism is the same. It is like shifting an inclined ruler from the origin
until it crosses one of the points.

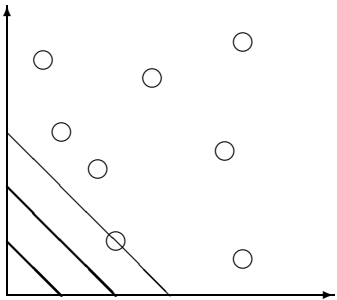
\setlength{\unitlength}{5000sp}%
\begingroup\makeatletter\ifx\SetFigFont\undefined%
\gdef\SetFigFont#1#2#3#4#5{%
  \reset@font\fontsize{#1}{#2pt}%
  \fontfamily{#3}\fontseries{#4}\fontshape{#5}%
  \selectfont}%
\fi\endgroup%
\begin{picture}(1674,1554)(4639,-4123)
\thinlines
{\color[rgb]{0,0,0}\put(4681,-4021){\vector( 0, 1){1440}}
\put(4681,-4021){\vector( 1, 0){1620}}
}%
{\color[rgb]{0,0,0}\put(5131,-3391){\circle{90}}
}%
{\color[rgb]{0,0,0}\put(5221,-3751){\circle{90}}
}%
{\color[rgb]{0,0,0}\put(4951,-3211){\circle{90}}
}%
{\color[rgb]{0,0,0}\put(4861,-2851){\circle{90}}
}%
{\color[rgb]{0,0,0}\put(5401,-2941){\circle{90}}
}%
{\color[rgb]{0,0,0}\put(5851,-2761){\circle{90}}
}%
{\color[rgb]{0,0,0}\put(5761,-3301){\circle{90}}
}%
{\color[rgb]{0,0,0}\put(5851,-3841){\circle{90}}
}%
\thicklines
{\color[rgb]{0,0,0}\put(4681,-3751){\line( 1,-1){270}}
}%
{\color[rgb]{0,0,0}\put(4681,-3481){\line( 1,-1){540}}
}%
\thinlines
{\color[rgb]{0,0,0}\put(4681,-3211){\line( 1,-1){810}}
}%
\end{picture}%
\nop{
|
|      o
|           o
|       o
|\           o
|  \                                    .
|    \        o
|      \                                .
|\       \                              .
|  \       \                            .
|    \       \        o
|      \       \                        .
+----------------------------
}

The slope of the line is determined by the weights. Arbitrary weights means
arbitrary slope. Merging takes the points that are first reached by shifting
the ruler, inclined arbitrarily.

The visual representation offers a simpler interpretation of this process: the
points first crossed by a line are the convex hull of the set of models; among
those, merging selects the models that are visible from the origin. It is like
the ideal point $[0,0]$ were a source of light, the models of $\mu$ formed a
convex figure, and merging selected the enlightened part of it. If either
formula is inconsistent with $\mu$, the figure is closed by lines parallel to
the axes, whose points are not selected.


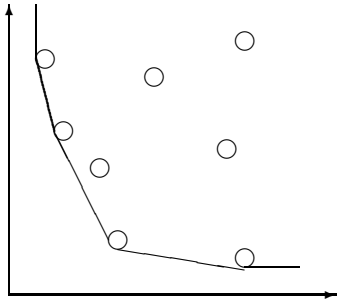
\setlength{\unitlength}{5000sp}%
\begingroup\makeatletter\ifx\SetFigFont\undefined%
\gdef\SetFigFont#1#2#3#4#5{%
  \reset@font\fontsize{#1}{#2pt}%
  \fontfamily{#3}\fontseries{#4}\fontshape{#5}%
  \selectfont}%
\fi\endgroup%
\begin{picture}(1674,1554)(4639,-4123)
\thinlines
{\color[rgb]{0,0,0}\put(4681,-4021){\vector( 0, 1){1440}}
\put(4681,-4021){\vector( 1, 0){1620}}
}%
{\color[rgb]{0,0,0}\put(5131,-3391){\circle{90}}
}%
{\color[rgb]{0,0,0}\put(5221,-3751){\circle{90}}
}%
{\color[rgb]{0,0,0}\put(4951,-3211){\circle{90}}
}%
{\color[rgb]{0,0,0}\put(4861,-2851){\circle{90}}
}%
{\color[rgb]{0,0,0}\put(5401,-2941){\circle{90}}
}%
{\color[rgb]{0,0,0}\put(5851,-2761){\circle{90}}
}%
{\color[rgb]{0,0,0}\put(5761,-3301){\circle{90}}
}%
{\color[rgb]{0,0,0}\put(5851,-3841){\circle{90}}
}%
{\color[rgb]{0,0,0}\put(5176,-3751){\line(-1, 2){270}}
\multiput(4906,-3211)(-0.96774,3.87097){94}{\makebox(0.7938,5.5563){\tiny.}}
\put(4816,-2851){\line( 0, 1){270}}
}%
{\color[rgb]{0,0,0}\put(5221,-3796){\line( 6,-1){627.568}}
\put(5851,-3886){\line( 1, 0){270}}
}%
\end{picture}%
\nop{
|      |
|      |
|      o
|      |    o
|       o
|         \  o
|           -
|             o
|               \                       .
|                 --
|                    \                  .
|                     o----------
|
+--------------------------------
}

This drawing not only explains the mechanism but also motivates it: rather than
rigidly evaluating the distances as shifting an inclined ruler over the plane,
merging selects the models that are qualitatively closer to the ideal condition
of zero distance.

\

The two counterexamples previously shown are easy to grasp when seen
graphically. The first proved that some non-dominated models are not minimal
using the three distance vectors $[0,3]$, $[2,2]$ and $[3,0]$. The second
vector is undominated, yet the line crossing the first and the third separates
it from the origin.

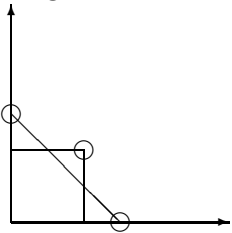
\setlength{\unitlength}{5000sp}%
\begingroup\makeatletter\ifx\SetFigFont\undefined%
\gdef\SetFigFont#1#2#3#4#5{%
  \reset@font\fontsize{#1}{#2pt}%
  \fontfamily{#3}\fontseries{#4}\fontshape{#5}%
  \selectfont}%
\fi\endgroup%
\begin{picture}(1145,1145)(3008,-3984)
{\color[rgb]{0,0,0}\thinlines
\put(3061,-3391){\circle{90}}
}%
{\color[rgb]{0,0,0}\put(3421,-3571){\circle{90}}
}%
{\color[rgb]{0,0,0}\put(3601,-3931){\circle{90}}
}%
{\color[rgb]{0,0,0}\put(3061,-3931){\vector( 0, 1){1080}}
\put(3061,-3931){\vector( 1, 0){1080}}
}%
{\color[rgb]{0,0,0}\put(3061,-3571){\line( 1, 0){360}}
\put(3421,-3571){\line( 0,-1){360}}
}%
{\color[rgb]{0,0,0}\put(3061,-3391){\line( 1,-1){540}}
}%
\end{picture}%
\nop{
  |
3 O
2 |\ O
  |  \           .
  +----O-
     2 3
}

The counterexample showing that two weight vectors may be necessary for
obtaining the correct result of merging is the converse, in which the central
point is closer to the origin than the line crossing the other two.

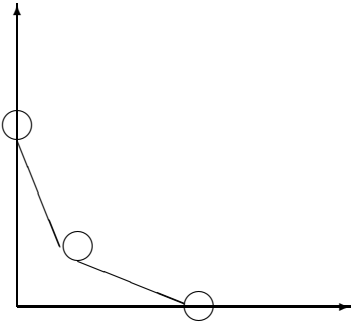
\setlength{\unitlength}{5000sp}%
\begingroup\makeatletter\ifx\SetFigFont\undefined%
\gdef\SetFigFont#1#2#3#4#5{%
  \reset@font\fontsize{#1}{#2pt}%
  \fontfamily{#3}\fontseries{#4}\fontshape{#5}%
  \selectfont}%
\fi\endgroup%
\begin{picture}(1745,1595)(4568,-4194)
{\color[rgb]{0,0,0}\thinlines
\put(4651,-3211){\circle{150}}
}%
{\color[rgb]{0,0,0}\put(4951,-3811){\circle{150}}
}%
{\color[rgb]{0,0,0}\put(5551,-4111){\circle{150}}
}%
{\color[rgb]{0,0,0}\put(4651,-4111){\vector( 0, 1){1500}}
\put(4651,-4111){\vector( 1, 0){1650}}
}%
{\color[rgb]{0,0,0}\put(4651,-3286){\line( 2,-5){212.069}}
}%
{\color[rgb]{0,0,0}\put(4951,-3886){\line( 5,-2){530.172}}
}%
\end{picture}%
\nop{
  |
3 O
  |
2 |\         .
  | |
1 | O
  |   \      .
  +-----O-
    1 2 3
}

All three models are enlightened from the origin (visible from it) and
therefore minimal. However, no single line crosses all of them. No single
weight vector selects all three.

The graphical representation also shows that two formulae have at most a linear
number of minimal distance vectors. Indeed, two of them cannot have the same
abscissa, as otherwise the one of larger ordinate would not be minimal. As a
result, every abscissa may have at most one minimal distance vector. Since the
points represent differing literals, the number of different abscissas is
linear. The number of weight vectors that are necessary to obtain the minimal
models is linear too.

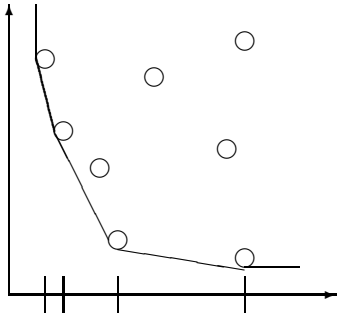
\setlength{\unitlength}{5000sp}%
\begingroup\makeatletter\ifx\SetFigFont\undefined%
\gdef\SetFigFont#1#2#3#4#5{%
  \reset@font\fontsize{#1}{#2pt}%
  \fontfamily{#3}\fontseries{#4}\fontshape{#5}%
  \selectfont}%
\fi\endgroup%
\begin{picture}(1674,1554)(4639,-4123)
\thinlines
{\color[rgb]{0,0,0}\put(4681,-4021){\vector( 0, 1){1440}}
\put(4681,-4021){\vector( 1, 0){1620}}
}%
{\color[rgb]{0,0,0}\put(5131,-3391){\circle{90}}
}%
{\color[rgb]{0,0,0}\put(5221,-3751){\circle{90}}
}%
{\color[rgb]{0,0,0}\put(4951,-3211){\circle{90}}
}%
{\color[rgb]{0,0,0}\put(4861,-2851){\circle{90}}
}%
{\color[rgb]{0,0,0}\put(5401,-2941){\circle{90}}
}%
{\color[rgb]{0,0,0}\put(5851,-2761){\circle{90}}
}%
{\color[rgb]{0,0,0}\put(5761,-3301){\circle{90}}
}%
{\color[rgb]{0,0,0}\put(5851,-3841){\circle{90}}
}%
{\color[rgb]{0,0,0}\put(5176,-3751){\line(-1, 2){270}}
\multiput(4906,-3211)(-0.96774,3.87097){94}{\makebox(0.7938,5.5563){\tiny.}}
\put(4816,-2851){\line( 0, 1){270}}
}%
{\color[rgb]{0,0,0}\put(5221,-3796){\line( 6,-1){627.568}}
\put(5851,-3886){\line( 1, 0){270}}
}%
{\color[rgb]{0,0,0}\put(4861,-3931){\line( 0,-1){180}}
}%
{\color[rgb]{0,0,0}\put(4951,-3931){\line( 0,-1){180}}
}%
{\color[rgb]{0,0,0}\put(5221,-3931){\line( 0,-1){180}}
}%
{\color[rgb]{0,0,0}\put(5851,-3931){\line( 0,-1){180}}
}%
\end{picture}%
\nop{
|      |
|      |
|      o
|      |    o
|       o
|         \  o
|           -
|             o
|               \                       .
|                 --
|                    \                  .
|                     o----------
|
+------||-----|-------|----------
}

\

Since the number of models is finite, the enlightened points can be determined
by standard geometric algorithms: quickhull~\cite{barb-etal-96} isolates the
segments that form the convex hull; the visible points are then singled
out~\cite{sech-gree-82}. These algorithms are efficient when the points are
given explicitly, but for the present case the points derive from propositional
models; therefore, they can be exponentially many in the number of variables
and in the size of formulae.

The pictorial representation of an undominated model suggests a mechanism for
establishing whether a model is undominated: for every other model, check
whether it falls in the rectangle having the origin and the model to check as
the opposite corners. This mechanism can be adapted to arbitrary weights and
two formulae by checking two other models at time instead of one. For $m$
formulae it checks groups of $m$ models each.

\begin{theorem}
\label{m-models}

For every distance $d$, formulae $\mu,F_1,\ldots,F_m$
and model $I \in \mod(\mu)$, it holds
{} $I \not\in \Delta^{d,W_\exists,\cdot}_\mu(F_1,\ldots,F_m)$
if and only if
{} $I \not\in \Delta^{d,W_\exists,\cdot}_{\mu'}(F_1,\ldots,F_m)$
for some formula $\mu'$ such that $I \models \mu'$, $\mu' \models \mu$ and
$|\mod(\mu')| \leq m+1$.

\end{theorem}

\proof If $I \not\in \Delta^{d,W_\exists,\cdot}_{\mu'}(F_1,\ldots,F_m)$ then
$I$ is minimal among the models of $\mu'$ for no weights $W \in W_\exists$.
Since $\mu' \models \mu$, all models of $\mu'$ are also models of $\mu$.
Therefore, $I$ is not minimal among the models of $\mu$ either. This proves one
direction of the claim, the rest of the proof is for the other.

The assumption is that
{} $I \not\in \Delta^{d,W_\exists,\cdot}_\mu(F_1,\ldots,F_m)$.
By definition, this condition holds if no weight vector $W=[w_1,\ldots,w_m]$
satisfies
{} $W > 0$ and
{} $W \cdot d(I,F_1,\ldots,F_m) \leq W \cdot d(J,F_1,\ldots,F_m)$
{}   for every $J \models \mu$.
This is a system of inequalities in variables $W=[w_1,\ldots,w_m]$. By
assumption, it is unfeasible. This condition is shown to be equivalent to the
existence of a formula $\mu'$ satisfied by $I$ and at most $m$ other models of
$\mu$ such that
{} $I \not\in \Delta^{d,W_\exists,\cdot}_{\mu'}(F_1,\ldots,F_m)$.

Every infeasible system of linear inequalities in $m$ variables that contains
more than $m+1$ is redundant: some of its inequalities are implied by the
others~\cite{chin-97}. This is an obvious consequence of a theorem by
Motzkin~\cite[Theorem~D1]{motz-52} on minimal unsatisfiable system of
inequalities, in modern terms called irredundant infeasible systems of
inequalities or IIS~\cite{amal-etal-99}. The theorem tells that every
irredundant infeasible system of $l$ inequalities has a coefficient matrix of
rank $l-1$. Since the rank is bounded by both the dimensions of the matrix, it
holds $l-1 \leq m$, meaning that $l \leq m+1$: an irredundant infeasible system
of inequalities in $m$ variables contains at most $m+1$ inequalities. The
original set of inequalities is unfeasible if and only if one of its subsets of
at most $m+1$ inequalities is infeasible. The theorem holds for non-strict
inequalities, but a simple change turns strict inequalities into strict
equalities while preserving both infeasibility and the coefficient
matrix~\cite[Section~67]{motz-52}\cite[Theorem~4]{gree-96}.

In the present case, the $k \leq m+1$ inequalities that form the infeasible and
irredundant subset of the original system of inequalities comprises some $w_i >
0$ and some $W \cdot d(I,F_1,\ldots,F_m) \leq W \cdot d(J,F_1,\ldots,F_m)$. The
second kind of inequalities are all satisfied by $W=[0,\ldots,0]$: a subset of
containing no inequality $w_i > 0$ is feasible. Therefore, at least one of the
$k \leq m+1$ inequalities is among $w_i>0$, and the inequalities of the form
{} $W \cdot d(I,F_1,\ldots,F_m) \leq W \cdot d(J,F_1,\ldots,F_m)${\plural}
are no more than $m$. If $J_1,\ldots,J_o$ are the models in the right-hand side
of these $o \leq m$ inequalities, the system comprising $W>0$ and $W \cdot
d(I,F_1,\ldots,F_m) \leq W \cdot d(J_i,F_1,\ldots,F_m)$ for $i=1,\ldots,o$ is
infeasible. This defines
{} $I \not\in \Delta^{d,W_\exists,\cdot}_{\mu'}(F_1,\ldots,F_m)$
where $\mu'$ is the formula satisfied exactly by $I$ and $J_1,\ldots,J_o$.~\qed

This theorem allows reformulating merging in terms of the groups of $m+1$
models of $\mu$.

\begin{theorem}
\label{intersection}

For every distance $d$ and formulae $\mu,F_1,\ldots,F_m$, it holds:

$$
\Delta^{d,W_\exists,\cdot}_\mu(F_1,\ldots,F_m) =
\bigcap_{|\mod(\mu')| \leq m+1, \mu' \models \mu}
\{ I \mid I \not\models \mu' \} \cup
\Delta^{d,W_\exists,\cdot}_{\mu'}(F_1,\ldots,F_m)
$$

\end{theorem}

\proof If $I \not\in \Delta^{d,W_\exists,\cdot}_\mu(F_1,\ldots,F_m)$
{} then $I \not\in \Delta^{d,W_\exists,\cdot}_{\mu'}(F_1,\ldots,F_m)$
for some $\mu'$ such that
{} $|\mod(\mu')| \leq m+1$, $\mu' \models \mu$ and $I \in \mu'$
by Theorem~\ref{m-models}.
{} 
Since $I$ is a model of $\mu'$, it does not belong to $\{ I \mid I \not\models
\mu' \}$ either. Therefore, it is not in
{} $\{ I \mid I \not\models \mu' \} \cup
{}  \Delta^{d,W_\exists,\cdot}_{\mu'}(F_1,\ldots,F_m)$.
Consequently, it is not in any intersection of this set with other sets. This
proves the first direction of the claim.

For the other direction, let $I$ be a model of $\mu$ that is not in
{} $ \bigcap_{|\mod(\mu')| \leq m+1, \mu' \models \mu}
{}   \{ I \mid I \not\models \mu' \} \cup
{}   \Delta^{d,W_\exists,\cdot}_{\mu'}(F_1,\ldots,F_m)$.
Since this is an intersection of sets, $I$ cannot belong to all the sets.
Let $\mu'$ be a formula such that
{} $I \not\in \{ I \mid I \not\models \mu' \} \cup
{}            \Delta^{d,W_\exists,\cdot}_{\mu'}(F_1,\ldots,F_m)$.
This is a union of sets, therefore $I$ does not belong to any:
{} $I \not\in \{I \mid I \not\models \mu'\}$ 
and
{} $I \not\in \Delta^{d,W_\exists,\cdot}_{\mu'}(F_1,\ldots,F_m)$.
The first condition implies that $I$ is a model of $\mu'$.
Since
{} $I \models \mu'$,
{} $|\mod(\mu')| \leq m+1, \mu' \models \mu$ and
{} $I \not\in \Delta^{d,W_\exists,\cdot}_{\mu'}(F_1,\ldots,F_m)$
Theorem~\ref{m-models} applies, proving that
{} $I \not\in \Delta^{d,W_\exists,\cdot}_\mu(F_1,\ldots,F_m)$.~\qed

Graphically, Theorem~\ref{m-models} is easier to grasp in terms of excluded
models rather than minimal models. A model is excluded (non-minimal) if and
only if there exists $m$ other models that exclude it: for every weight vector
one of these models has lower distance than its. With two formulae, each pair
of models defines an excluded area; every model that they exclude is not
minimal. Therefore, merging is the result of over-imposing all these excluded
areas.

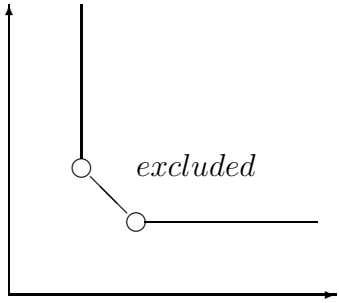
\setlength{\unitlength}{5000sp}%
\begingroup\makeatletter\ifx\SetFigFont\undefined%
\gdef\SetFigFont#1#2#3#4#5{%
  \reset@font\fontsize{#1}{#2pt}%
  \fontfamily{#3}\fontseries{#4}\fontshape{#5}%
  \selectfont}%
\fi\endgroup%
\begin{picture}(1674,1494)(4549,-4153)
{\color[rgb]{0,0,0}\thinlines
\put(4951,-3481){\circle{90}}
}%
{\color[rgb]{0,0,0}\put(5221,-3751){\circle{90}}
}%
{\color[rgb]{0,0,0}\put(4591,-4111){\vector( 0, 1){1440}}
\put(4591,-4111){\vector( 1, 0){1620}}
}%
{\color[rgb]{0,0,0}\put(4996,-3526){\line( 1,-1){180}}
}%
{\color[rgb]{0,0,0}\put(4951,-3436){\line( 0, 1){765}}
}%
{\color[rgb]{0,0,0}\put(5266,-3751){\line( 1, 0){855}}
}%
\put(5221,-3526){\makebox(0,0)[lb]{\smash{{\SetFigFont{12}{24.0}{\rmdefault}{\mddefault}{\updefault}{\color[rgb]{0,0,0}$excluded$}%
}}}}
\end{picture}%
\nop{
|    |
|    |
|    o   excluded
|      \                  .
|        o------
|
+------------------
}

The region that is upper-right of the broken line is the area of exclusion of
the two models (the horizontal and vertical half-lines are also excluded, but
not the diagonal segment). Each pair of models cuts out a similar area from the
plane if none dominate the other. The result of deleting all of them is the
result of merging.

Whether a model is in one such area is determined algebraically. The points in
it are separated from the origin by any one of the three lines of this figure:

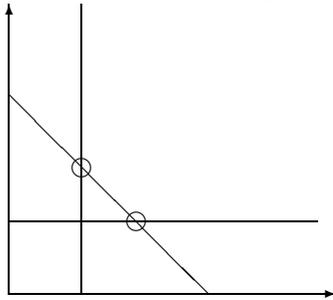
\setlength{\unitlength}{5000sp}%
\begingroup\makeatletter\ifx\SetFigFont\undefined%
\gdef\SetFigFont#1#2#3#4#5{%
  \reset@font\fontsize{#1}{#2pt}%
  \fontfamily{#3}\fontseries{#4}\fontshape{#5}%
  \selectfont}%
\fi\endgroup%
\begin{picture}(1674,1494)(4549,-4153)
{\color[rgb]{0,0,0}\thinlines
\put(4951,-3481){\circle{90}}
}%
{\color[rgb]{0,0,0}\put(5221,-3751){\circle{90}}
}%
{\color[rgb]{0,0,0}\put(4591,-4111){\vector( 0, 1){1440}}
\put(4591,-4111){\vector( 1, 0){1620}}
}%
{\color[rgb]{0,0,0}\put(4951,-2671){\line( 0,-1){1440}}
}%
{\color[rgb]{0,0,0}\put(4591,-3751){\line( 1, 0){1530}}
}%
{\color[rgb]{0,0,0}\put(4591,-3121){\line( 1,-1){990}}
}%
\end{picture}%
\nop{
|\   |
|  \ |
|    o
|    | \	       .
|----+---o------
|    |     \           .
+------------------
}

Checking whether a line separates a point from the origin is trivial: given the
equation of the line $ax+by+c=0$, the expression $ax+by+c$ is zero on the
points in the line and none other. By setting $x$ and $y$ with same sign of $a$
and $b$ respectively and sufficiently large ($abs(x) > abs(c/a)$ if $a \neq 0$,
the same for $y$), the expression has positive value. With the opposite signs,
it has negative values. Since the function is continuous and is only zero on
the line, it has the same sign in each half-plane. As a result, a point is
separated from the origin if $ax+by+c$ and $a0+b0+c$ have different signs. The
same condition can also be checked by computing the sign of two angles among
the involved points~\cite{petr-13}.

That the three lines define the area of exclusion of two models is now formally
proved. Let $D$, $D'$ and $D''$ be the distance vectors of three models. The
first is excluded by the other two if the following system of inequalities in
two variables $W=[w_1,w_2]$ has no solution.

\begin{eqnarray*}
w_1 &>& 0				\\
w_2 &>& 0				\\
W \cdot D &\leq& W \cdot D'		\\
W \cdot D &\leq& W \cdot D''
\end{eqnarray*}

These are linear inequalities; therefore, they hold in a polygon. Each segment
of its perimeter is part of a line; this line is an extreme point for the
inequalities, and is therefore obtained by converting some of these
inequalities into equations. Every line corresponds to a single pair of weights
$W=[w_1,w_2]$, as each such pair defines a line in the plan.

The first conversion turns
{} $w_1 > 0$ and $W \cdot D \leq W \cdot D'$ into
{} $w_1=0$ and $W \cdot D = W \cdot D'$.
These equations imply $w_2 d_2 = w_2 d_2'$, which is the same as $d_2=d_2'$
since $w_2$ is not null. This equation defines the horizontal line that crosses
$D'$. A similar restriction leads to the horizontal line crossing $D''$. The
highest of these two lines is irrelevant, since the half-plane over it is all
contained in that of the other. What remain is a horizontal line.

In the same way, converting $w_2 > 0$ into $w_2=0$ leads to the vertical line.

Converting both $w_1>0$ and $w_2>0$ into $w_1=0$ and $w_2=0$ does not lead to a
line, since all points satisfy the remaining equalities in this case.

The only conversion left to try is to turn
{} $W \cdot D \leq W \cdot D'$ and $W \cdot D \leq W \cdot D''$
into
{} $W \cdot D = W \cdot D'$ and $W \cdot D = W \cdot D''$.
These equations imply $W \cdot D' = W \cdot D''$. If $D'$ strictly dominates
$D''$ this is a contradiction, and no line is generated; the same if $D''$
strictly dominates $D'$. Otherwise, $W \cdot D' = W \cdot D''$ implies that $W
\cdot D = W \cdot D'$ is the same as $W \cdot D = W \cdot D''$: these two
equations with variables $D$ are equivalent. They are the same linear equation
in two variables; therefore, they define a line. Since $D=D'$ and $D=D''$ both
satisfy them, this line crosses the points $D'$ and $D''$.

\

An alternative condition for the exclusion of a model from merging is based on
the lines separating other models from the origin. In particular: $I \models
\mu$ is excluded if either $d(J,F_1,F_2) < d(I,F_1,F_2)$ for some model $J$ of
$\mu$, or two other models $J$ and $K$ of $\mu$ are such that:

\begin{enumerate}

\item $d(J,F_1,F_2) \not< d(K,F_1,F_2)$;

\item $d(K,F_1,F_2) \not< d(J,F_1,F_2)$;

\item the line crossing $I$ and $J$ does not separate $K$ from the origin; and

\item the line crossing $I$ and $K$ does not separate $J$ from the origin.

\end{enumerate}

Rather than proving the claim, the region of models $I$ such that the line
crossing each and $K$ separates $J$ from the origin is shown on the figure for
two models $J$ and $K$ that do not dominate each other.

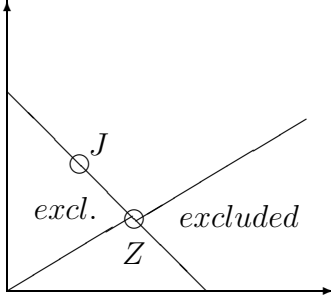
\setlength{\unitlength}{5000sp}%
\begingroup\makeatletter\ifx\SetFigFont\undefined%
\gdef\SetFigFont#1#2#3#4#5{%
  \reset@font\fontsize{#1}{#2pt}%
  \fontfamily{#3}\fontseries{#4}\fontshape{#5}%
  \selectfont}%
\fi\endgroup%
\begin{picture}(1674,1494)(4549,-4153)
{\color[rgb]{0,0,0}\thinlines
\put(4951,-3481){\circle{90}}
}%
{\color[rgb]{0,0,0}\put(5221,-3751){\circle{90}}
}%
{\color[rgb]{0,0,0}\put(4591,-4111){\vector( 0, 1){1440}}
\put(4591,-4111){\vector( 1, 0){1620}}
}%
{\color[rgb]{0,0,0}\put(5221,-3751){\line(-1, 1){630}}
\put(4591,-3121){\line( 0,-1){990}}
\put(4591,-4111){\line( 5, 3){622.059}}
}%
{\color[rgb]{0,0,0}\put(6076,-3256){\line(-5,-3){847.059}}
\put(5221,-3751){\line( 1,-1){360}}
\put(5581,-4111){\line( 1, 0){495}}
}%
\put(4996,-3436){\makebox(0,0)[lb]{\smash{{\SetFigFont{12}{24.0}{\rmdefault}{\mddefault}{\updefault}{\color[rgb]{0,0,0}$J$}%
}}}}
\put(5221,-3976){\makebox(0,0)[b]{\smash{{\SetFigFont{12}{24.0}{\rmdefault}{\mddefault}{\updefault}{\color[rgb]{0,0,0}$Z$}%
}}}}
\put(5041,-3751){\makebox(0,0)[rb]{\smash{{\SetFigFont{12}{24.0}{\rmdefault}{\mddefault}{\updefault}{\color[rgb]{0,0,0}$excl.$}%
}}}}
\put(5446,-3796){\makebox(0,0)[lb]{\smash{{\SetFigFont{12}{24.0}{\rmdefault}{\mddefault}{\updefault}{\color[rgb]{0,0,0}$excluded$}%
}}}}
\end{picture}%
\nop{
|     
|..                
|....J           ..
|......\     ......
|........K.........
|...       ........
+------------------
}

Adding the models in the same condition with $J$ and removing the models that
are strictly dominated by either $J$ or $K$ leaves the correct area.

\

These conditions allow for an algorithm that progressively excludes models.
Initially, all models of $\mu$ are provisionally accepted. Then, a single model
is chosen and the models it strictly dominates removed. The procedure continues
on the remaining models until nothing changes. At this point, no model
dominates another. A pair of models is selected; the conditions above determine
whether this pair excludes some other models, which are removed. When the
procedure ends, only the models selected by merging are left.

{\bf Algorithm~1:} $\Delta^{d,W_\exists,\cdot}_\mu(F_1,F_2)$

\begin{enumerate}

\item $M = \mod(\mu)$;

\item $C = \true$

\item while $C$ do:

\begin{enumerate}

\item $C = \false$

\item for each $I,J \in M$

\begin{itemize}

\item[] if $d(J,F_1,F_2) < d(I,F_1,F_2)$ then

\begin{enumerate}

\item $M = M \backslash \{I\}$

\item $C = \true$

\end{enumerate}

\end{itemize}

\end{enumerate}

\item $C = \true$

\item while $C$ do:

\begin{enumerate}

\item $C = \false$

\item for each $I,J,K \in M$

\begin{itemize}

\item[] if
the line crossing $d(I,F_1,F_2)$ and $d(K,F_1,F_2)$
does not separate $J$ from the origin and
the line crossing $d(I,F_1,F_2)$ and $d(K,F_1,F_2)$
does not separate $K$ from the origin
then

\begin{enumerate}

\item $M = M \backslash \{I\}$

\item $C = \true$

\end{enumerate}

\end{itemize}

\end{enumerate}

\item return $M$

\end{enumerate}

The algorithm first removes all models that are strictly dominated others. The
removed models can be forgotten, since they are not in the part of the convex
hull that is visible from the origin; therefore, they are unnecessary for
excluding other models. When all strictly dominated models are out from $M$,
the algorithm proceeds by comparing each model $I$ with a pair of other models
$J$ and $K$. These two do not strictly dominate each other because all models
that are strictly dominated have already been removed from $M$. This is why no
dominance check is done in Step~5b.

\section{Postulates}

Merging with unknown weights depends on the distance $d$ and the set of weight
vectors $W.$. Some postulates for belief merging hold for all sets of weights
vectors (IC0-IC3 and IC7), others only for some including $W_\exists$
(IC4-IC6), and one does not hold for $W_\exists$ (IC8). Merging with completely
unknown weights $W_\exists$ cannot be expressed as a preorder, not even a
partial one.

\

Postulates IC0-8~\cite{koni-pere-11} cannot all hold, since a merging operator
satisfying all of them can be expressed as a selection of models of $\mu$ that
are minimal according to some total preorder depending only on
$F_1,\ldots,F_m$. Actually, not even a partial preorder expresses merging with
unknown weights.

\begin{theorem}
\label{no-preorder}

No partial preorder $\leq$ depending on $F_1$ and $F_2$ only is such that
$\Delta^{dh,W_\exists,\cdot}_\mu(F_1,F_2) = \min(\mod(\mu),\leq)$.

\end{theorem}

\proof By Lemma~\ref{distances-formulae}, for every set of distance vectors
there exists $\mu$, $F_1$ and $F_2$ such that the models of $\mu$ have these
Hamming distance vectors from $F_1$ and $F_2$. The distance vectors that prove
the claim are $[3,0]$, $[2,2]$ and $[0,3]$.

Let $\mu'$ be the formula having only the models at distance $[2,2]$ and
$[3,0]$. They are both minimal with weights $W=[2,1]$. As a result, $[3,0]
\not< [2,2]$. The formula $\mu''$ with the models at distance $[2,2]$ and
$[0,3]$ shows that $[0,3] \not< [2,2]$. The ordering $\leq$ is the same since
by assumption it does not depend on $\mu$ but only on $F_1$ and $F_2$, which
are the same. A consequence of $[3,0] \not< [2,2]$ and $[0,3] \not< [2,2]$ is
that the model at distance $[2,2]$ is minimal among the three. However, it is
not in the result of merge with constraints $\mu$ as proved by
Lemma~\ref{undominated-excluded}.~\qed

The line of proof reveals which postulate is not satisfied: removing a model
from $\mu$ is the same as conjoining $\mu$ with another formula; the postulate
forbidding the inclusion of new models is IC8. The proof shows that IC8 does
not hold.

How about the others? Some hold for every set of distance vectors $W.$, others
only for some. Some postulates hold only if the distance $d$ satisfies the
triangle inequality, others hold even if $d(I,F)$ is not defined in terms of a
distance among models $d(I,J)$. The latter requires $d(I,F) \in Z_{\geq 0}$ and
$d(I,F)=0$ if and only if $I \models F$. In the following summary, this case is
described as ``a model-formula distance''. In this section, $E$ is sometimes
used in place of $F_1,\ldots,F_m$ following the notation by Konieczny and
Perez~\cite{koni-pere-11}. This simplifies some formulae.


\begin{enumerate}
\setcounter{enumi}{-1}

\item
$\Delta^{d,W.,\cdot}_\mu(E) \models \mu$

holds for every model-formula distance and non-empty set of weight vectors

\item if $\mu$ is consistent, then $\Delta_\mu(E)$ is consistent

holds for every model-formula distance and non-empty set of weight vectors

\item if $\bigwedge E$ is consistent with $\mu$,
then $\Delta^{d,W.,\cdot}_\mu(E) = \mu \wedge \bigwedge E$

holds for every model-formula distance and non-empty set of weight vectors

\item if $E_1 \equiv E_2$ and $\mu_1 \equiv \mu_2$, then
$\Delta^{d,W.,\cdot}_{\mu_1}(E_1) \equiv \Delta^{d,W.,\cdot}_{\mu_2}(E_2)$

holds for every model-formula distance and non-empty set of weight vectors

\item
if $F_1 \models \mu$ and $F_2 \models \mu$ then
{} $\Delta^{d,W.,\cdot}_\mu(F_1,F_2) \wedge F_1$ is consistent if and only if
{} $\Delta^{d,W.,\cdot}_\mu(F_1,F_2) \wedge F_2$ is consistent.

holds if $W.$ contains every permutation of every vector it contains
($W_\exists$ has this property, as well as $W_a$ for every $a \in \Z_{>0}$) and
$d$ satisfies the triangle inequality: $d(I,K) + d(K,J) \geq d(I,J)$ (both $dd$
and $dh$ have this property); if any of these two conditions do not hold, a
counterexample shows that the postulate does not hold

\item
{} $\Delta^{d,W.',\cdot}_\mu(F_1,\ldots,F_k) \wedge
{}  \Delta^{d,W.'',\cdot}_\mu(F_{k+1},\ldots,F_m)
{} \models
{} \Delta^{d,W.,\cdot}_\mu(F_1,\ldots,F_m)$

requires $W. = W.' \times W.''$

\item
if
{} $\Delta^{d,W.',\cdot}_\mu(F_1,\ldots,F_k) \wedge
{}  \Delta^{d,W.'',\cdot}_\mu(F_{k+1},\ldots,F_m)$
is consistent, it is entailed by
{} $\Delta^{d,W.,\cdot}_\mu(F_1,\ldots,F_m)$

requires $W. = W.' \times W.''$

\item $\mu' \wedge \Delta^{d,W.,\cdot}_\mu(E) \models
       \Delta^{d,W.,\cdot}_{\mu \wedge \mu'}(E)$

holds for every model-formula distance and non-empty set of weight vectors

\item if $\mu' \wedge \Delta^{d,W.,\cdot}_\mu(E)$ is consistent, then
$\Delta^{d,W.,\cdot}_{\mu \wedge \mu'}(E) \models \Delta^{d,W.,\cdot}_\mu(E)$

does not hold for the Hamming distance $dh$ and the set of all weight vectors
$W_\exists$

\end{enumerate}

The formal proofs of these claims follow. First, postulates IC0-IC3 and IC7
hold for every non-empty set of weight vectors $W.$ and model-formula distance.

\begin{lemma}

For every model-formula distance $d(I,F)$ and non-empty set of weight vectors
$W.$, the merging operator $\Delta^{d,W.,\cdot}$ satisfies postulates IC0-IC3
and IC7.

\end{lemma}

\proof The claim is proved one postulate at time.

\begin{enumerate}
\setcounter{enumi}{-1}

\item $\Delta^{d,W.,\cdot}_\mu(E) \models \mu$

by definition, $\Delta^{d,W.,\cdot}_\mu(E)$ is a subset of the models of
$\mu$;

\item if $\mu$ is consistent, then $\Delta_\mu(E)$ is consistent

by assumption, $W.$ contains at least a vector of weights $W$; for this vector,
$\Delta^{d,W,\cdot}_\mu(E)$ is the set of models of $\mu$ at minimal weighted
distance from $F_1,\ldots,F_m$; if $\mu$ is consistent, it has at least a
minimal model;

\item if $\bigwedge E$ is consistent with $\mu$,
then $\Delta^{d,W.,\cdot}_\mu(E) = \mu \wedge \bigwedge E$

since $d(I,F_i)=0$ when $I \models F_i$, the distance vectors of the models of
$\bigwedge E$ are $[0,\ldots,0]$; regardless of the weights, the weighted
distance is zero, and therefore minimal; all other models have a strictly
positive distance; since weights are strictly positive, their weighted distance
is greater than zero;

\item if $E_1 \equiv E_2$ and $\mu_1 \equiv \mu_2$, then
$\Delta^{d,W.,\cdot}_{\mu_1}(E_1) \equiv \Delta^{d,W.,\cdot}_{\mu_2}(E_2)$

equivalence between profiles $E_1 \equiv E_2$ holds when there exists a
bijection such that the associated formulae are equivalent; this postulate
holds because merging is defined from the models of the formulae;

\setcounter{enumi}{6}

\item $\mu' \wedge \Delta^{d,W.,\cdot}_\mu(E) \models
       \Delta^{d,W.,\cdot}_{\mu \wedge \mu'}(E)$

the models of $\mu' \wedge \Delta^{d,W.,\cdot}_\mu(E)$, if any, are the models
that satisfy $\mu'$, and also satisfy $\mu$ and no other model of $\mu$ has a
lower distance from $E$ weighted by some $W \in W.$; each such model satisfies
$\mu \wedge \mu'$, and no other model of $\mu \wedge \mu'$ has lower distance
weighted by $W$, since the models of $\mu \wedge \mu'$ are a subset of those of
$\mu$;

\end{enumerate}

\qed

This lemma does not even require $d(I,F)$ to be defined in terms of a distance
between models. The next one does, and additionally needs the triangle
inequality.

\begin{lemma}

If $W.$ contains every permutation of every vector it contains and $d$
satisfies the triangle inequality $\forall I,J,K d(I,K) + d(K,J) \geq d(I,J)$,
then IC4 holds. For some set of weight vectors that does not include a
permutation of one of its elements IC4 does not hold. The same for some
distance not satisfying the triangle inequality.

\end{lemma}

\proof Postulate 4 is: if $F_1 \models \mu$ and $F_2 \models \mu$ then
{} $\Delta^{d,W.,\cdot}_\mu(F_1,F_2) \wedge F_1$ is consistent if and only if
{} $\Delta^{d,W.,\cdot}_\mu(F_1,F_2) \wedge F_2$ is consistent.

This postulate does not hold in general. For example, the single weight vector
$[2,1]$ with the drastic or Hamming distance and $\mu=\true$, $F_1=x$ and
$F_2=\neg x$ would select the model of $F_1$ only. Both distances satisfy the
triangle inequality.

This counterexample suggests that the postulate holds if the set $W.$ has some
sort of symmetry: if it contains a weight vector, it also contains all its
permutations. This is not however not sufficient, as shown by the following
counterexample:

\begin{eqnarray*}
ds(I,J) &=&
\left\{
\begin{array}{ll}
0 & \mbox{if } dh(I,J) = 0			\\
1 & \mbox{if } dh(I,J) = 1			\\
2 & \mbox{if } 2 \leq dh(I,J) \leq 4		\\
5 & \mbox{if } 5 \leq dh(I,J)
\end{array}
\right.									\\
W. &=& \{[5,2],[2,5]\}							\\
F_1 &=&
x_1 \wedge x_2 \wedge x_3 \wedge x_4 \wedge x_5				\\
F_2 &=&
\neg x_1 \wedge \neg x_2 \wedge \neg x_3 \wedge \neg x_4 \wedge \neg x_5\\
\mu &=&
F_1 \vee F_2 \vee
(x_1 \wedge \neg x_2 \wedge \neg x_3 \wedge \neg x_4 \wedge \neg x_5)
\end{eqnarray*}

The distance $ds$ may look unnatural, but has a rationale: instead of measuring
the distance between models by the exact number of differing literals, it
roughly approximates it by aggregating certain groups of consecutive values
into one, so that only a finite number of possible distances would result.

The models of $\mu$ have distance vectors $[0,5]$, $[2,1]$ and $[5,0]$. The
first and the third are the models of $F_1$ and $F_2$, respectively. The weight
vector $[5,2]$ turns these distance vectors into the weighted distances
{} $[5,2] \cdot [0,5] = 10$,
{} $[5,2] \cdot [2,1] = 12$,
{} $[5,2] \cdot [5,0] = 25$;
only the model of $F_1$ is minimal. For the weight vector $[2,5]$:
{} $[2,5] \cdot [0,5] = 25$,
{} $[2,5] \cdot [2,1] = 9$,
{} $[2,5] \cdot [5,0] = 10$;
the only minimal model is the second, which is not a model of $F_2$. This is a
case in which both $F_1$ and $F_2$ imply $\mu$ and
$\Delta^{ds,W.,\cdot}_\mu(F_1,F_2)$ is consistent with $F_1$ but not with
$F_2$.

Note that Lemma~\ref{distances-formulae} does not apply to this case. It tells
how to obtain certain distance vectors with formulae $\mu,F_1,F_2$, but these
do not necessarily obey $F_1 \models \mu$ and $F_2 \models \mu$. To the
contrary, the proof itself shows that both $F_1$ and $F_2$ have models that
falsify $\mu$.

\

Postulate 4 requires not only $W.$ to be symmetric, but also $d$ to satisfy the
triangle inequality: for every three models $I$, $J$ and $K$ it holds $d(I,K) +
d(K,J) \leq d(I,J)$.

Since $\Delta^{d,W.,\cdot}_\mu(F_1,F_2) \wedge F_1$ is consistent, there
exists a weight vector $[a,b]$ and a model $I \in \models F_1$ with distance
vector $[0,c]$ such that $[a,b] \cdot [0,c]$ is minimal (the zero is because
$I \models F_1$).

By definition, $d(I,F_2) = c$ means that $d(I,J) = c$ for some $J \in
\mod(F_2)$. This implies that $d(J,F_1) \leq d(J,I) = c$; if $d(J,F_1) < c$
then $d(J,K) < c$ for some $K \in \mod(F_1)$, which implies that $d(K,F_2) < c
= d(I,F_2)$, contradicting the assumption that $I$ is minimal; therefore,
$d(J,F_1) = c$.

Since $J$ satisfies $F_2$, it also satisfies $\mu$. It is therefore a candidate
for being in the result of merge. If $a < b$, then the weighted distance of $J$
would be $[a,b] \cdot [c,0] = a \times c < b \times c = [a,b] \cdot [0,c]$.
Since $[0,c]$ is the distance vector of $I$, this contradicts the assumption
that $I$ is minimal for weights $[a,b]$. This proves that $a \geq b$.

Model $J$ is now proved to have minimal distance weighted by $[b,a]$. The
weighted distance of $J$ is $[b,a] \cdot [c,0] = b \times c$. Contrary to the
claim, let $K$ be a model with distance vector $[e,f]$ such that $[b,a] \cdot
[e,f] < b \cdot c$.

The triangular property implies $e+f \geq c$. In details: $e+f<c$ implies the
existence of $I'$ and $J'$ of respectively $F_1$ and $F_2$ such that
$d(K,I')=e$, $d(K,J')=f$ and $d(I',J') \leq e+f < c$. This contradicts the
assumption of minimality of $I$. This property $e+f \geq c$, together with $a
\geq b$, makes the following inequalities valid:

\begin{eqnarray*}
[b,a] \cdot [e,f]
&=&
b \times e + a \times f				\\
&=&
b \times e + b \times f + (a-b) \times f	\\
&=&
b \times (e + f) + (a-b) \times f		\\
&\geq&
b \times c + (a-b) \times f			\\
&\geq&
b \times c
\end{eqnarray*}

Contrary to what assumed $[b,a] \cdot [e,f] \geq b \times c$. This proves that
no such model $K$ may exists, and that $J$ has minimal distance weighted by
$[b,a]$. Since $J \models F_2$, the required consistency of
{} $\Delta^{d,W.,\cdot}_\mu(F_1,F_2) \wedge F_2$ is proved.~\qed

Since $W_\exists$ is symmetric and both $dd$ and $dh$ satisfy the triangle
inequality, Postulate~4 holds in these two cases. Actually, for $W_\exists$ the
distance does not matter, and $\Delta^{d,W_\exists,\cdot}(F_1,\ldots,F_m)$ is
always consistent with every $F_i$ that is consistent with $\mu$. If the
maximal value of the distance from $F_1$ and from $F_2$ is $k$, the weight
vectors $[k+1,1]$ and $[1,k+1]$ suffice. The first guarantees that every model
of $F_1$ is always better than one of $\neg F_1$, no matter how close the
second is to $F_2$. The same for the second weight vector.

This lemma shows an effect of the triangle inequality on belief merging. It is
a quite natural requirement and is obeyed by both the drastic and the Hamming
distance, but is mostly useless in belief merging~\cite{koni-pere-11}; so far
it only seemed to affect the infinite-alphabet case~\cite{chac-pere-04} and the
application of belief revision to case-based reasoning~\cite{coja-lieb-12}.

Postulates 5 and 6 require special care to be even formulated. Informally, they
tell that merging $F_1,\ldots,F_k,F_{k+1},\ldots,F_m$ is the same as merging
$F_1,\ldots,F_k$, then merging $F_{k+1},\ldots,F_m$ and finally conjoining the
two results if they do not conflict. This is simple to express if no weights
are involved, otherwise each of these three integrations is defined over its set of
weights. If these are unrelated, like $[1,\ldots,1,1\ldots,1]$ for the overall
merge and $[10,1,\ldots,1]$ and $[1,\ldots,10]$ for merging the two parts, the
three results cannot be expected to be coherent.

This is why Postulates 5 and 6 cannot be said to be obeyed plain and simple.
Rather, they are satisfied only when the sets of weights are related in the
appropriate way.

\begin{lemma}

If
{} $\Delta^{d,W.',\cdot}_\mu(F_1,\ldots,F_k) \wedge
{}  \Delta^{d,W.'',\cdot}_\mu(F_{k+1},\ldots,F_m)$
is consistent, it is equivalent to
{} $\Delta^{d,W.,\cdot}_\mu(F_1,\ldots,F_m)$,
where $W. = W.' \times W.''$ (postulates IC5 and IC6).

\end{lemma}

\proof Let $I$ be a model of
{} $\Delta^{d,W.',\cdot}_\mu(F_1,\ldots,F_k) \wedge
{}  \Delta^{d,W.'',\cdot}_\mu(F_{k+1},\ldots,F_m)$.
By assumption, there exists $W' \in W.'$ and $W'' \in W.''$ such that the
distance vector $d(I,F_1,\ldots,F_k)$ weighted by $W'$ is minimal among the
models of $\mu$, and the distance vector $d(I,F_{k+1},\ldots,F_m)$ weighted by
$W''$ is minimal among the models of $\mu$. This is equivalent to
$d(I,F_1,\ldots,F_k,F_{k+1},\ldots,F_m)$ being minimal when weighted by
$W'W''$; this is the vector obtained by concatenating $W'$ and $W''$, and is
therefore in $W. = W.' \times W.''$.

In the other way around, a model that is not minimal on its weighted distance
to $F_1,\ldots,F_k$ or to $F_{k+1},\ldots,F_m$ is not minimal on its weighted
distance to $F_1,\ldots,F_k,F_{k+1},\ldots,F_m$.~\qed

IC8 does not hold. The following counterexample shows that for the Hamming
distance $dh$ and the set of all weight vectors $W_\exists$.

\begin{theorem}

There exists
$\mu$, $\mu'$, $F_1$ and $F_2$ such that 
{} $\mu' \wedge \Delta^{dh,W_\exists,\cdot}_\mu(E)$ is consistent
but
{} $\Delta^{dh,W_\exists,\cdot}_{\mu \wedge \mu'}(E)
{}  \not\models
{}  \Delta^{dh,W_\exists,\cdot}_\mu(E)$.

\end{theorem}

\proof Let $\mu$, $F_1$ and $F_2$ be such that $\mu$ has three models with
distance vectors
{} $d(I,F_1,F_2) = [1,0]$,
{} $d(J,F_1,F_2) = [0,1]$ and
{} $d(K,F_1,F_2) = [0,2]$.
Such formulae exist thanks to Lemma~\ref{distances-formulae}.

The models of $\Delta^{dh,W.,\cdot}_\mu(F_1,F_2)$ are $I$ and $J$, since these
two models have minimal distance weighted by $[1,1]$. Since $J$ dominates $K$,
by Lemma~\ref{dominance-always} $K$ is not in the result of merge for any
weights.

Let $\mu'$ be the formula with models $I$ and $K$. Since $I$ also satisfies
$\Delta^{dh,W_\exists,\cdot}_\mu(F_1,F_2)$, the conjunction of this and $\mu'$
is consistent. When merging under constraints $\mu \wedge \mu'$, model $K$ is
now minimal with weights $[2,1]$, since it and the other model $I$ of $\mu
\wedge \mu'$ have both weighted distance $3$.~\qed

This counterexample completes the analysis of the basic postulates IC0-IC8. Two
additional ones exist: majority and arbitration. The first tells that a formula
repeated enough times is entailed by the result of merging; the second was
initially defined as the irrelevance of the number of repetitions, and has a
newer definition that is difficult to summarize in words.

Majority does not hold with $W_\exists$. Not that it should. No matter how many
times a formula is repeated, regardless of how many sources supports it, its
negation may come from a single source that is more reliable than all the
others together. When reliability is uncertain, this case has to be taken into
account. It is not even uncommon in practice: many commonly held belief are in
fact false.

Many commonly held belief are in fact false: Napoleon was short~\cite{duna-63};
diamonds had been typical gemstones for engagement rings since a long
time~\cite{epst-82}, the red telephone is a telephone line, and one of its end
is in the White House~\cite{clav-13}; meteorites are always hot when they reach
the Earth's surface; flowering sunflowers turn to follow the sun (only the gems
do); the Nazis issued an ultimatum before the Ardeatine massacre (something
even witnesses of the time believe)~\cite[p.~155]{mazz-03}; fans in closed
rooms kill people (many people in Korea believed this). A page on Wikipedia
list more than a hundred of commonly believed facts that are in fact
false~\cite{wiki-misc-17}. The material was enough for a 26-episodes TV
show~\cite{wiki-adam-17}.

All of this shows that no matter how many times a fact is repeated, if the
reliability of its sources is unsure it may still be falsified by a single
reliable source. This is what the following theorem formally proves.

\begin{theorem}

There exists $F_1,F_2$ such that
$\Delta^{d,W_\exists,\cdot}_\true(F_1,F_2,\ldots,F_2) \not\models F_2$, where
$F_2$ is repeated an arbitrary number of times.

\end{theorem}

\proof The formulae are $F_1=a$ and $F_2=\neg a$. For every number of
repetitions $n$, there exists $W$ such that
$\Delta^{d,W,\cdot}_\mu(F_1,F_2,\ldots,F_2)$ contains the model $\{a\}$, which
does not satisfy $F_2$. In particular, the weights are $W=[n,1,\ldots,1]$. The
weighted distance of $\{a\}$ from the formulae is $n$, the same as the weighted
distance of the only other model $\{\neg a\}$. As a result, $\{a\}$ is
minimal.~\qed

Arbitration was initially defined as the opposite condition of irrelevance of
the number of repetitions~\cite{koni-pere-98,meye-01}. This property holds for
$W_\exists$. The following theorem proves an equivalent formulation of it.

\begin{lemma}

For every $\mu,F_1,\ldots,F_m$ it holds:

$$
\Delta^{d,W_\exists,\cdot}_\mu(F_1,\ldots,F_m)
=
\Delta^{d,W_\exists,\cdot}_\mu(F_1,\ldots,F_m,F_m)
$$

\end{lemma}

\proof By definition,
{} $\Delta^{d,W_\exists,\cdot}_\mu(F_1,\ldots,F_m)$
is the union of
{} $\Delta^{d,W,\cdot}_\mu(F_1,\ldots,F_m)$
for every $W \in W_\exists$, and the same for merging with a duplicated
$F_m$. The claim is proved by showing that for each
{} $W=[w_1,\ldots,w_{m-1},w_m]$
there exists
{} $W'=[w_1',\ldots,w_{m-1}',w_m',w_m'']$
such that
{} $\Delta^{d,W,\cdot}_\mu(F_1,\ldots,F_m)$ is equal to
{} $\Delta^{d,W',\cdot}_\mu(F_1,\ldots,F_m,F_m)$, and vice versa.

The distance of a model from $F_1,\ldots,F_m$ weighted by
{} $W=[w_1,\ldots,w_{m-1},w_m]$
is exactly half of the distance of the same model from $F_1,\ldots,F_m,F_m$
weighted by
{} $W'=[2 \times w_1,\ldots,2 \times w_{m-1},w_m,w_m]$,
since each distance is multiplied by two. Therefore, the minimal models are the
same.

Vice versa, the distance of a model from $F_1,\ldots,F_m,F_m$ weighted by
{} $W'=[w_1,\ldots,w_{m-1},w_m,w'_m]$
is exactly the same as the distance of the same model from $F_1,\ldots,F_m$
weighted by
{} $W=[w_1,\ldots,w_{m-1},w_m+w'_m]$.
In this case, the weighted distances are exactly the same, and the minimal
models coincide.~\qed

A newer version of the arbitration postulate is expressed in terms of the
preorder between models as:
{} if $I<_{F_1}J$, $I<_{F_2}J'$ and $J \equiv_{F_1,F_2} J'$
{} then $I <_{F_1,F_2} J$.
As proved by Theorem~\ref{no-preorder}, merging under unknown reliability
cannot be expressed as a preorder, total or otherwise. The expression of the
postulate in terms of formulae is even more convoluted, and is not clear
whether it makes sense when merging is not expressible in terms of a preorder.

\section{The disjunctive property}

A further condition merging may or may not meet is the disjunctive property,
defined on two formulae as Postulate~7 by Liberatore and
Schaerf~\cite{libe-scha-98-b} and later generalized to an arbitrary number of
formulae with integrity constraints by Everaere et al.~\cite{ever-etal-10}. In
terms of models it has a simple intuitive expression. Every model is a possible
state of the world; merging only selects worlds that at least one of the
sources consider possible. In formulae, a model $I$ is in the result of merging
only if $I \models F_i$ for at least one of the merged formulae $F_i$. Since
$I$ must also satisfy the integrity constraints $\mu$, this requirement is
lifted when none of the formulae $F_i$ is consistent with $\mu$.

This condition is not satisfied by $\Delta^{dh,W_=,\cdot}$; as a result, is not
satisfied by $\Delta^{dh,W_\exists,\cdot}$ either since $W_= \subset
W_\exists$. However, a suitable set of weight vectors allows for a specific
form of disjunctive merging, that based on closest pairs of models.

\begin{definition}[\cite{libe-scha-98-b}]

Merging by closest pairs of models is defined from the ordering between pairs
of models
{} $\l I,J \r \leq_dh \l I',J' \r$ if and only if $dh(I,J) \leq dh(I',J')$
by selecting the models in all minimal pairs:

$$
F_1 \Delta_D F_2 = \{
	I,J
	\mid
	\l I,J \r \in \min(\mod(F_1) \times \mod(F_2), \leq_dh) \}
$$

\end{definition}

The set of weight vectors used for obtaining this definition is
{} $W_{n+1} = \{[1,n+1], [n+1,1]\}$,
the specific form of $W_a$ when $a$ is the number of variables increased by one
and merging is between two formulae.

\begin{theorem}

For every pair of satisfiable formulae $F_1$ and $F_2$ over an alphabet of $n$
variables, it holds
{} $F_1 \Delta_D F_2 = \Delta^{dh,W_{n+1},\cdot}_\true(F_1,F_2)$.

\end{theorem}

\proof By definition, $I \in F_1 \Delta_D F_2$ if and only $I \models F_1$ and
there exists $J \models F_2$ such that $\l I,J \r$ is minimal according to
$\leq_dh$, or the same with $F_1$ and $F_2$ swapped. What is
now proved is that the first condition is equivalent to
{} $I \in \Delta^{dh,[n+1,1],\cdot}_\true(F_1,F_2)$.
By symmetry, the condition with the two formulae swapped is equivalent to
{} $I \in \Delta^{dh,[1,n+1],\cdot}_\true(F_1,F_2)$
without the need of a proof.

The relevant cases are:
{} $I \models F_1$
{}   and $\l I,J \r$ is minimal for some $J \models F_2$,
{} $I \models F_1$
{}   and $\l I,J \r$ is minimal for no $J \models F_2$,
{} and $I \not\models F_1$.
The claim holds if $I \in \Delta^{dh,W_{n+1},\cdot}_\true(F_1,F_2)$ holds
exactly in the first case.

\begin{enumerate}

\item
{} $I \models F_1$
{}   and $\l I,J \r$ is minimal for some $J \models F_2$;
since $I \in F_1$, the distance from $I$ to $F_1$ is zero: $dh(I,F_1) = 0$;
therefore, the weighted distance from $I$ to the formulae is $[n+1,1] \cdot
[0, dh(I,F_2)] = dh(I,F_2)$, which is at most $n$; the negation of the claim is
that the weighted distance $(n+1) \times dh(K,F_1) + 1 \times dh(K,F_2)$ of
some other model $K$ is less than that; for it being less than $n$ implies
$dh(K,F_1)=0$; as a result, the weighted distance of $K$ is $dh(K,F_2)$; if it
were less than the weighted distance of $I$ then $dh(K,F_2) < dh(I,F_2)$; by
definition, this means that there exists $K'$ such that $dh(K,K')$ is less than
$dh(I,I')$ for every $I' \models F_2$, including $I' = J$; this means that
$dh(K,K') < dh(I,J)$, contrary to the assumption that $\l I,J \r$ is minimal;

\item
{} $I \models F_1$
{}   and $\l I,J \r$ is minimal for no $J \models F_2$;
by assumption, there exists $K,K'$ such that $K \models F_1$, $K' \models F_2$
and $dh(K,K') < dh(I,J)$ for every $J \models F_2$; this means that $dh(K,F_2)
< dh(I,F_2)$; since both $I$ and $K$ satisfy $F_1$, it also holds $dh(I,F_1) =
dh(K,F_1) = 0$; as a result, the weighted distances of these models are
{} $[n+1,1] \cdot [0,dh(I,F_2)] = dh(I,F_2)$ and
{} $[n+1,1] \cdot [0,dh(K,F_2)] = dh(K,F_2)$;
since $dh(K,F_2) < dh(I,F_2)$, the model $I$ is not at a minimal weighted
distance;

\item
{} $I \not\models F_1$;
since $F_1$ is by assumption satisfiable, it has a model $K$; since
$dh(K,F_1)=0$, the weighted distance for this model is
{} $[n+1,1] \cdot [dh(K,F_1),dh(K,F_2)] = dh(K,F_2)$,
which is at most $n$; the weighted distance of $I$ is instead
{} $[n+1,1] \cdot [dh(I,F_1),dh(K,F_2)] = (n+1) \times dh(I,F_1) + dh(I,F_2)$,
which is greater than $n$ since $dh(I,F_1)>0$.

\end{enumerate}

Since $I$ has minimal weighted distance from $F_1$ and $F_2$ in the first case
but not in the second and the third, the claim is proved.~\qed

A disjunctive operator on $m$ formulae is obtained similarly when all formulae
are consistent and the integrity constraints are void: $\mu=\true$.

\begin{theorem}

For every distance $d$ bounded by $k$ and such that $d(I,F)=0$ if and only if
$I \models F$, if $F_1,\ldots,F_m$ are satisfiable then
{} $\Delta^{d,W_{k \times m},\cdot}_\true(F_1,\ldots,F_m)$
is a disjunctive merging operator.

\end{theorem}

\proof Let $I$ be a model satisfying no formula $F_i$. The disjunctive property
holds if $I$ is not in
{} $\Delta^{d,W_{k \times m},\cdot}_\true(F_1,\ldots,F_m)$.
This holds if $I$ is not in
{} $\Delta^{d,W,\cdot}_\true(F_1,\ldots,F_m)$
for any $W \in W_{k \times m}$.

By assumption, $I$ does not satisfy any of the formulae. Therefore, its
distance vector is less than or equal to $[1,\ldots,1]$. Multiplying this
vector by $W$, the result is $k \times m + (m-1)$.

Let $i$ be the index such that the $i$-th element of $W_{k \times m}$ is $k
\times m$. Since $F_i$ is satisfiable, it has a model $J$. The distance vector
of $J$ is at most $[k,\ldots,k,0,k,\ldots,k]$ where $0$ is at index $i$. The
result of multiplying it by $W$ is $(m-1) \times k$.

The upper bound for the weighted distance of $J$ is $(m-1) \times k$, which is
less than $k \times m + (m-1)$, the lower bound of the weighted
distance of minimal $I$. This proves that $I$ is not minimal.~\qed

The set of weight vectors used in this and the previous theorem provides an
alternative view of the disjunctive property. One line of reasoning when
merging conflicting information is to assume that a single source $i$ is
completely right; consequently, the information that one provides is accepted
in full, and is integrated with that coming from the other sources only as long
as this is consistent with it. Since the actual reliability is unknown, no
unique choice of $i$ is warranted. Technically, every formula that is
consistent with the integrity constraints could come from the completely
reliable source; the weight vectors $W_a$ for a sufficiently large $a$
formalize this mechanism.

This is a correct way of implementing the principle of indifference in belief
merging: rather than assuming that all sources are equally reliable, one of
them is taken as completely right, but this is done for each of them at time.
Indifference is realized by symmetry, not equality.

\section{Sources providing multiple formulae}

The previous sections are about combining a number of independent formulae.
This is the basic problem of belief merging: each formula comes from a
different source, so that their reliabilities are independent. This is
formalized by the weights being unconstrained in the set $W_\exists$.

When a source provides more than one formula, each of them is as reliable as
its source. The same mechanism employing the weighted sum of the drastic or
Hamming distance can be used, but the weights are associated to the sources
rather than to the formulae. All formulae from the same source have the same
reliability and therefore the same weight (this condition is close in spirit to
the unit partitions by Booth and Hunter~\cite{boot-hunt-18}).

Technically, each source is represented by a set of formulae $S_i$. Its
reliability is encoded by a positive integer $w_i$. Given a set
$\{S_1,\ldots,S_m\}$ of such sources, merging is done by selecting the minimal
models of the integrity constraints $\mu$ according to this evaluation:


\ttytex{
$$
v(I) = \sum_{S_i} w_i \times \sum{F_i \in S_i} d(I,F_i)
$$
}{
v(I) = SUM(Si) wi x SUM(Fi in Si) d(I,Fi)
}

This is the $\mathrm{DA}^2$ operator~\cite{koni-lang-marq-04} with the sum as
intra-source aggregation and the weighted sum as the inter-source aggregation.
The sum is subject to the problem of manipulation: a source may provide the
same formula multiple times in order to influence the final
result~\cite{chop-etal-06}; this is a problem especially when merging
preferences, but not when merging beliefs with unknown reliability. Even if a
source provides the same formula a thousand times, one of the considered
alternatives is that the weight of this source is a thousand times smaller than
the others, making such a manipulation ineffective.

The only technical result in this section is that merging with the drastic
distance is not the same as disjoining the maxcons. This is proved by the
following sources with $\mu=\true$.

\begin{eqnarray*}
S_1 &=&	\{x,y,z\}			\\
S_2 &=&	\{\neg x,\neg y\}		\\
S_3 &=&	\{\neg x,\neg z\}
\end{eqnarray*}

One of the maxcons of $\{x,y,z,\neg x,\neg y,\neg z\}$ is $\{x,\neg y,\neg
z\}$, which is not obtained when merging with unknown reliabilities.
Intuitively, to include the formula $x$ from $S_1$ in the result that formula
needs to count at least twice as much as each formula $\neg x$ from $S_2$ and
$S_3$, but this implies the same for $y$ and $z$, which excludes $\neg y$ and
$\neg z$.

Formally, let the weights of the sources be $w_1,w_2,w_3$. The weighted
distance of some relevant models are:

\begin{eqnarray*}
I = \{x,\neg y,\neg z\}
& ~ &
v(I) = w_1 \times 2 + w_2 \times 1 + w_3 \times 1 = w_1 \times 2 + w_2 + w_3
\\
J = \{\neg x,\neg y,\neg z\}
& ~ &
v(J) = w_1 \times 3 + w_2 \times 0 + w_3 \times 0 = w_1 \times 3
\\
K = \{x,y,z\}
& ~ & 
v(K) = w_1 \times 0 + w_2 \times 2 + w_3 \times 2 = w_2 \times 2 + w_3 \times 2
\end{eqnarray*}

In order for $I$ to be minimal $v(I)$ must be less than or equal to $v(J)$ and
$v(K)$:

\begin{eqnarray*}
w_1 \times 2 + w_2 + w_3	& \leq &	w_1 \times 3		\\
w_1 \times 2 + w_2 + w_3	& \leq &	w_2 \times 2 + w_3 \times 2
\end{eqnarray*}

This system of inequalities is infeasible. The first implies $w_2 + w_3 \leq
w_1$, which makes the right-hand side of the second become less than or equal
to $w_1 \times 2$, while it should instead be greater than or equal than $w_1
\times 2 + w_2 + w_3$, and therefore greater than $w_1 \times 2$. This proves
that $\{x,\neg y,\neg z\}$ is not a minimal model for any weight vector.

A similar example shows that merging does not result in the maxcons of the
conjunctions of the sources. Let $S_1=\{x,y\}$ and $S_2=\{\neg x,\neg y\}$. The
only maxcons of $\{\wedge S_1, \wedge S_2\}$ are $x \wedge y$ and $\neg x
\wedge \neg y$, but merging with unknown weights selects all four models over
$x$ and $y$, since this is the result when $W=[1,1]$.

\section{Conclusions}

Sometimes the information to be merged comes from sources of equal reliability.
In such cases, merging with equal weights is correct. But if reliability is
unknown, assuming weights equal is unwarranted. The difference is not only
conceptual but also technical. Theorem~\ref{no-preorder} shows that merging
with unknown reliability cannot in general be reduced to a preorder among
models, not even a partial one.

A result emerged by the study of this setting is a motivation for merging by
maxcons~\cite{bara-etal-92}. This mechanism has sometimes been considered
unsuitable for merging because it disregards the distribution of information
among sources~\cite{koni-00,koni-pere-11}. Theorem~\ref{drastic-maxcons} shows
it the same as merging with the drastic distance when reliability is unknown.
The number of repetitions of a formula is irrelevant to this kind of
merging---as it should. A formula only occurring once may come from a very
reliable source, while its negation is supported only by untrustable sources.
Without any knowledge of the reliability of the sources, this is a situation to
take into account.

This article not only backs merging by maxcons, but more generally merging by
the (MI) postulate: the number of repetitions of a formula is irrelevant to
merging. Of course, there are many cases where this postulate should not hold;
whenever reliability is known, two sources providing a formula give twice the
support for it. But when reliability is unknown every numeric evaluation
becomes irrelevant, including doubling the support for a formula like in this
case. As already discussed by Meyer~\cite{meye-01}, Postulate (MI) may
sometimes be right; it is inconsistent with the other postulates IC0-IC8, but
the fault is on them. While Meyer blames Postulate~4 of merging without
integrity constraints~\cite{koni-pere-98}, merging with unknown reliability
conflicts with IC8.

A minor technical contribution of this article is a case for the distance
function to obey the triangle inequality. This property had only a couple of
applications in belief revision and merging so
far~\cite{chac-pere-04,coja-lieb-12}, but is generally not
required~\cite{koni-pere-11}. The new consequence of it shown in this article
is that it allows satisfying Postulate (IC4) when merging with unknown
reliability.

\

Some problems of merging with unknown reliability are left open.

The graphical representation of merging by a convex hull has been shown to work
in the bidimensional case, which corresponds to merging two formulae only. It
looks like it works for an arbitrary number of formulae, but this is still left
to be proved.

Another open problem is the characterization of merging with unknown
reliability when sources provide more than one formula each. This is the case
treated by the $\mathrm{DA}^2$ merging operators~\cite{koni-lang-marq-04}. It
has been only touched in the previous section, but that preliminary analysis
already shows that this case cannot in general be reduced to that of
one-formula sources.

While the weighed sum is one of the basic mechanism for
merging~\cite{reve-97,lin-96,lin-mend-99,koni-lang-marq-04}, it is not the only
one. The Max~\cite{reve-97}, Gmax~\cite{koni-pere-02} and quota
merging~\cite{ever-etal-10} semantics are alternatives to be used under
different conditions; for example, Gmax is the system to use when sources are
assumed to be unlikely to be very far from truth.

\

A comparison with related work follows.

The weighted distance from a set of formulae was first used for merging by
Revesz~\cite{reve-97}, and investigated by Lin~\cite{lin-96}. Lin and
Mendelzon~\cite{lin-mend-99} and Konieczny and Perez~\cite{koni-pere-98}
used the unweighted sum for merging. These articles assume either equal or
fixed weights, not varying weights like is done in the present one.


Benferhat, Lagrue and Rossit~\cite{benf-etal-14} considered the related problem
of commensurability: when the sources themselves assess the reliability of the
formulae they provide, they may not use the same scale; this is related to a
similar issue in social choice theory. Their study and the present one differ
in formalism (ranked bases instead of formulae with distance functions), but
they share the principle of considering a set of alternative reliability
assessments. There is however an early point of departure: Benferhat, Lagrue
and Rossit~\cite{benf-etal-14} distill a single preorder and then select the
models that are minimal according to it; Theorem~\ref{no-preorder} shows that
the same cannot be done in general in the settings of the present article. A
point of contact is the case of drastic distance: Theorem~\ref{binary} could be
alternatively proved from certain results by Benferhat, Lagrue and
Rossit~\cite[Propositions 1,2,8]{benf-etal-14}.


Accepting what is true according to all possible relative reliabilities is
analogous to drawing the consequences that hold in all probability measures in a
set~\cite{halp-tutt-93}, and can be seen as the formal logic version of the
``worst scenario'' in economics: ``{\em the firm may not be certain about the
``relative plausibility'' of these boom probabilities. [...] if the firm acts
in accordance with certain sensible axioms, then its behavior can be
characterized as being uncertainty-averse: when the firm evaluates its
position, it will use a probability corresponding to the ``worst''
scenario}''~\cite{nish-ozak-07}. Belief revision and merging aim at the most
knowledge that can be justifiably and consistently obtained; therefore, minimal
knowledge takes the place of the least profit, and the worst scenario for a
formula is one where it is false.







\sloppy
\bibliographystyle{plain}

\end{document}